# Enhancing Personalized Bladder Cancer Treatment Through Reinforcement Learning: A Recurrent Patient State-Transition Decision Support Framework

**Divyansh Chawla[1] , Anshu Garg[1], Isshaan Singh[1]**

[1] School of Computer Science and Engineering, Vellore Institute of Technology, Chennai-600 0127, Tamil Nadu, INDIA

Corresponding author: Isshaan Singh (Email id: Isshaan.singh2021@vitstudent.ac.in).

**ABSTRACT** Bladder cancer treatment requires personalized and adaptive decision-making, particularly in recurrent disease management, where treatment effectiveness evolves over successive clinical episodes. Conventional clinical decision support systems often rely on static treatment guidelines or single-step predictive models, limiting their ability to capture the longitudinal dynamics of disease progression and recurrence. This paper presents a recurrent patient state-transition simulation framework for bladder cancer treatment planning that integrates predictive state-transition modeling within a Markov Decision Process (MDP) and a Deep Q-Network (DQN) reinforcement learning environment. The predictive module estimates changes in tumor characteristics following treatment interventions, while the reinforcement learning agent sequentially optimizes treatment decisions by interacting with simulated patient trajectories. This modular framework enables dynamic, patient-specific treatment planning by continuously adapting recommendations based on evolving clinical states rather than relying solely on static prediction. Furthermore, the framework provides interpretable treatment trajectories and comprehensive simulation logs that support clinical decision-making by improving transparency and explaining the rationale underlying sequential treatment recommendations. To evaluate the proposed framework, its learning characteristics were compared with existing reinforcement learning-based treatment planning approaches. The proposed framework achieved a cumulative reward of 63,918.87, an average training loss per episode of 0.0056, and a policy improvement score of 6.62%, demonstrating effective sequential learning and robust treatment optimization within the simulated recurrent treatment environment. These findings highlight the potential of recurrent patient state-transition simulation combined with reinforcement learning as a flexible and extensible decision-support methodology for personalized bladder cancer treatment planning and future AI-assisted precision oncology.

## I. INTRODUCTION

The need for a robust decision support system (DSS) in healthcare, particularly for cancer treatment, has become increasingly vital as personalized medicine evolves. In the context of bladder cancer, the complexity and variability of treatment responses among patients highlight the importance of developing systems that tailor interventions based on individual patient data. Traditional cancer treatment methods often rely on generalized protocols, which fail to account for the dynamic and unique nature of each patient's condition [1]. Therefore, an intelligent system capable of analyzing diverse patient-specific factors and generating personalized treatment recommendations is crucial.

By incorporating advanced computational models such as machine learning and RL, it is possible to simulate treatment outcomes and assist healthcare providers in making more informed, precise decisions that improve patient outcomes while minimizing adverse effects [2][3]. Significant progress has been made in using data-driven models for cancer treatment planning. Many existing systems utilize machine learning techniques such as classification models, regression analysis, and statistical modeling to predict treatment outcomes based on historical patient data [28]. While these systems have shown promising results in various cancer types, challenges remain in integrating real-time patient data, adapting to new information, and ensuring that recommendations align with clinical decision-making processes [4]. Moreover, the application of RL in oncology has gained traction, with studies exploring its potential to optimize treatment plans over time, dynamically adjusting recommendations based on patient responses and evolving conditions. However, gaps remain in developing systems that seamlessly integrate clinical knowledge, patient preferences, and real-time data into a decision support framework that is both accurate and interpretable for healthcare providers [5].

### A. RESEARCH GAP

Despite the growing use of machine learning and RL in healthcare, existing DSS for bladder cancer treatment still face several limitations. Most current models rely on static predictive approaches that fail to dynamically adapt to evolving patient conditions. Additionally, while RL has shown promise in oncology, existing studies lack seamless integration with real-time clinical workflows, limiting their practical applicability. Furthermore, there is a gap in ensuring that RL-based treatment recommendations align with clinical guidelines while remaining interpretable and actionable for healthcare professionals. Addressing these gaps requires a framework that combines predictive modeling with adaptive decision-making, continuously refining treatment recommendations based on patient responses.

### B. MOTIVATION

Bladder cancer treatment is highly patient-specific, with significant variations in disease progression, treatment response, and comorbidities. A personalized and adaptive approach is essential to optimizing treatment outcomes and minimizing adverse effects. A framework is needed for recurrent event modeling in the context of bladder cancer, where treatment strategies must dynamically adapt to disease progression and patient responses. The motivation behind this research is to bridge the gap between predictive modeling and reinforcement learning (RL) in oncology by developing a DSS that continuously learns The motivation behind this research is to bridge the gap between predictive modeling and RL in oncology by developing a DSS that continuously learns from historical and real-time data. By integrating ML with RL, we aim to create a robust framework capable of providing adaptive, personalized treatment plans that evolve based on patient-specific metrics. This approach aligns with the broader goal of advancing AI-driven precision medicine, ensuring that treatment recommendations are not only data-driven but also clinically relevant and interpretable.

### C. KEY CONTRIBUTIONS

The Contributions of this research are as follows:

- A specialized environment tailored for bladder cancer treatment is developed. This environment simulates treatment analysis, allowing the agent to determine optimal treatment plans. By focusing on bladder cancer, the system is grounded in real-world clinical challenges, ensuring that recommendations are practical, relevant, and actionable.
- A novel approach combining traditional predictive models with RL is introduced. The RL agent learns from historical data, continuously refining its treatment predictions based on feedback and reward formulations. As the agent learns, it dynamically adjusts its treatment suggestions, ensuring that recommendations evolve over time and become increasingly accurate. This approach marks a significant advancement in AI-driven decision support systems in clinical oncology, providing tailored, personalized treatment recommendations based on the unique needs of each patient.

In this paper, we propose a DSS designed for deployment as an intelligent agent within a clinical environment, specifically aimed at optimizing bladder cancer treatment. The platform will integrate with electronic health records (EHRs), clinical databases, and treatment guidelines to provide real-time recommendations. By utilizing RL, the agent will continuously refine its decision-making process based on treatment outcomes, ensuring that patient care evolves dynamically. The system's primary function is to predict the most suitable treatment at each stage of a patient's journey, allowing for personalized and adaptive treatment planning.

A crucial aspect of developing an effective DSS for cancer treatment is data preprocessing. Clinical datasets, including EHRs, genomic information, and imaging data, often contain missing values, noise, and inconsistencies, necessitating robust preprocessing techniques [6,7]. Methods such as data imputation, feature scaling, and dimensionality reduction are essential to enhance the quality of input data [8-10]. Additionally, integrating multi-modal data such as genomic profiles and clinical parameters requires careful feature selection and alignment to ensure accurate predictions. The effectiveness of preprocessing directly impacts the performance of machine learning and RL models, making it a foundational step in system development.

To enhance predictive accuracy, classification techniques and deep learning models are incorporated into the system. By combining these classification methods with RL, the system establishes an adaptive learning environment where the agent refines its recommendations based on patient responses, clinical feedback, and predefined reward functions. This integration enables the agent to not only predict optimal treatment options but also continuously improve its recommendations over time through iterative learning.

## II. RELATED WORKS

Several studies have explored the use of EHRs to evaluate treatment efficacy, employing both statistical and machine learning models to assess treatment responses. These studies reveal significant heterogeneity among patient populations, emphasizing the need for personalized treatment approaches. Many research efforts have integrated clinical and genomic data to enhance predictive accuracy, while others have leveraged deep learning and Bayesian models for tailored treatment recommendations. Furthermore, large-scale EHR datasets have provided real-world evidence, uncovering patterns in treatment outcomes and highlighting the necessity of incorporating patient demographics, comorbidities, and treatment sequencing into predictive models [11]. Building on these insights, our study develops a robust statistical framework to analyze treatment heterogeneity, offering personalized treatment estimates to improve patient care and decision-making.

Existing healthcare systems have been designed to enhance patient care through the integration of advanced decision-support tools and the redesign of care pathways. These systems incorporate patient preferences, real-world evidence, and quality-of-life metrics to support personalized treatment strategies. Artificial intelligence (AI) and data-driven insights have been employed to improve shared decision-making between clinicians and patients. A key feature of such systems is the development of predictive models that assess treatment outcomes based on clinical, genetic, and lifestyle factors. Additionally, the integration of multidisciplinary care fosters a holistic approach to patient management, with real-time data analysis enhancing patient engagement and improving healthcare outcomes [12].

The CLARIFY platform is an advanced AI-driven system designed to analyze multidimensional cancer data for precise risk stratification and personalized follow-up care. It integrates diverse data sources, including EHRs, imaging, genomic profiles, and patient-reported outcomes, to generate comprehensive insights. Using machine learning algorithms, CLARIFY identifies high-risk patients, predicts disease progression, and recommends tailored interventions. A distinguishing feature of this platform is its adaptability to individual patient characteristics, ensuring accurate treatment planning. Additionally, it facilitates real-time monitoring and early detection of complications, enhancing long-term patient management. By providing evidence-based decision support, CLARIFY optimizes treatment pathways, ultimately improving cancer care efficiency and patient quality of life [13].

While DSS in oncology have improved clinical workflows and adherence to evidence-based practices, their impact on patient survival rates remains inconsistent. This underscores the complexity of cancer treatment, where factors such as tumor heterogeneity, patient comorbidities, and resistance to therapies play crucial roles. Although DSS can optimize treatment decisions and reduce medical errors, their effectiveness is often limited by challenges such as data integration, clinician trust, and real-world applicability. Future enhancements incorporating AI-driven predictive models and personalized medicine strategies may unlock greater benefits in patient outcomes [14].

Mathematical and machine learning models have been increasingly employed to predict tumor growth and simulate treatment outcomes, significantly enhancing personalized care. These models integrate compartmental approaches, such as logistic and Gompertz models, with machine learning algorithms like random forests and neural networks. By analyzing clinical, genomic, and imaging data, these models refine predictions and optimize treatment strategies. Personalizing treatment plans based on patient-specific characteristics ensures more targeted and effective cancer care, ultimately improving patient outcomes [15].

Advanced probabilistic models, including Bayesian approaches, have been utilized to determine optimal treatment pathways in oncology. The Oncology Simulation Model (OSM) framework, for instance, employs a continuous dynamic Markov cohort model to estimate disease prevalence and survival rates. By simulating treatment outcomes and adapting to evolving guidelines, these models support data-driven decision-making in oncology, ensuring optimized patient care and resource allocation [16].

Machine learning techniques, such as artificial neural networks (ANNs) and support vector machines (SVMs), have been widely used to predict treatment effectiveness and patient outcomes. These models analyze complex datasets, identifying key factors influencing treatment response and survival rates. ANNs are particularly effective in capturing non-linear patterns, while SVMs excel in classifying treatment responses. By integrating clinical, genetic, and imaging data, these models enable more accurate and personalized treatment strategies [17].

Multiscale models that incorporate tumor-immune interactions and resource competition have been developed to optimize chemotherapy regimens. These models simulate complex biological processes, identifying optimal dosing schedules that maximize treatment efficacy while minimizing adverse effects. Aligning with clinical data ensures that model predictions are realistic and applicable to real-world patient care [18]. AI technologies in oncology have facilitated personalized treatment selection and monitoring through clinical decision support systems and virtual biopsies. However, concerns regarding data transparency and bias remain significant challenges. While AI-driven models enhance treatment precision, ongoing research is necessary to ensure fairness, interpretability, and equitable application in clinical practice [19].

RL techniques, particularly those using Markov Decision Processes (MDPs), have been employed in cancer treatment planning to optimize therapeutic pathways. Multi-objective deep RL approaches have demonstrated effectiveness in balancing chemotherapy dosing to minimize side effects while maximizing treatment efficacy. By continuously adapting to patient-specific data, RL models enhance personalized treatment strategies, improving both survival rates and quality of life [20].

Fuzzy RL models combine fuzzy logic with reinforcement learning to address uncertainty in patient data and treatment responses. By incorporating flexible decision-making frameworks, these models optimize drug dosing schedules, reducing adverse effects and improving treatment precision. This hybrid approach enhances patient safety while improving therapeutic outcomes [21]. Despite advancements, RL models in healthcare face challenges such as data sparsity and lack of transparency. The "black-box" nature of many RL models limits their adoption in clinical settings, as clinicians require interpretability to trust AI-driven recommendations. Ensuring explainable AI (XAI) solutions will be critical for integrating RL into real-world medical decision-making [22][23].
Recent research has demonstrated the potential of RL models in predicting tumor recurrence and progression in bladder cancer. By integrating genomic and radiomic data, these models offer improved predictive accuracy, enabling more tailored treatment strategies. Continuous learning and real-time adaptation allow these models to refine treatment recommendations dynamically, improving patient outcomes and survival rates. As data availability and model transparency improve, RL-based approaches could play a transformative role in bladder cancer treatment [25][26].

The review highlights the significance of integrating AI and reinforcement learning into decision support systems for cancer treatment. Existing approaches leverage machine learning models, probabilistic frameworks, and RL techniques to optimize therapeutic strategies. Despite progress, challenges such as data integration, clinical applicability, and model interpretability persist. The proposed system addresses these limitations by incorporating adaptive RL-based decision-making with real-time patient data, ensuring personalized and continuously evolving treatment recommendations. By enhancing model transparency and integrating clinical workflows, this approach aims to bridge the gap between AI-driven insights and practical oncology applications, ultimately improving bladder cancer patient outcomes.

## III. METHODOLOGY

In our methodology, we formulate bladder cancer treatment optimization as a Markov Decision Process (MDP) by constructing a data-driven patient state-transition simulation framework capable of modeling disease progression across recurrent treatment episodes. Given the recurrent nature of bladder cancer, where treatment efficacy evolves over successive interventions, accurately representing longitudinal recurrence patterns is essential for informed sequential decision-making. Our dataset captures key clinical parameters from bladder cancer treatment records, including the initial size of the largest tumor, the number of tumors, treatment duration, and treatment history. Three treatment types—placebo, pyridoxine, and thiotepa—are administered over multiple treatment episodes, enabling the framework to model changes in tumor characteristics while preserving the temporal sequence of patient interventions.

Within the MDP framework, states represent the patient's clinical condition, encompassing tumor size, tumor count, treatment duration, and the recurrence count of administered treatments, while treatments constitute the available actions that govern transitions between successive patient states. To accurately represent disease progression over time, an enumeration feature is incorporated to distinguish individual treatment episodes by tracking the treatment sequence for each patient, thereby facilitating recurrent event modeling throughout the simulation process.

To emulate patient-specific disease progression within this sequential environment, a predictive state-transition module is learned from historical clinical observations. In the current implementation, Random Forest regressors are employed to estimate changes in tumor size and tumor count, allowing the framework to approximate subsequent patient states following each treatment decision. Rather than serving as the primary contribution, these predictive models function as transition approximators within the simulation environment, enabling repeated evaluation of treatment trajectories under different clinical scenarios. The simulated patient trajectories provide the reinforcement learning agent with an interactive environment in which treatment strategies can be iteratively evaluated and refined to optimize sequential decision-making. Cumulative changes in tumor size and tumor count serve as performance indicators, guiding the agent toward treatment policies that minimize long-term tumor progression. Figure 1 presents an overview of the proposed decision-support framework and the overall study workflow.

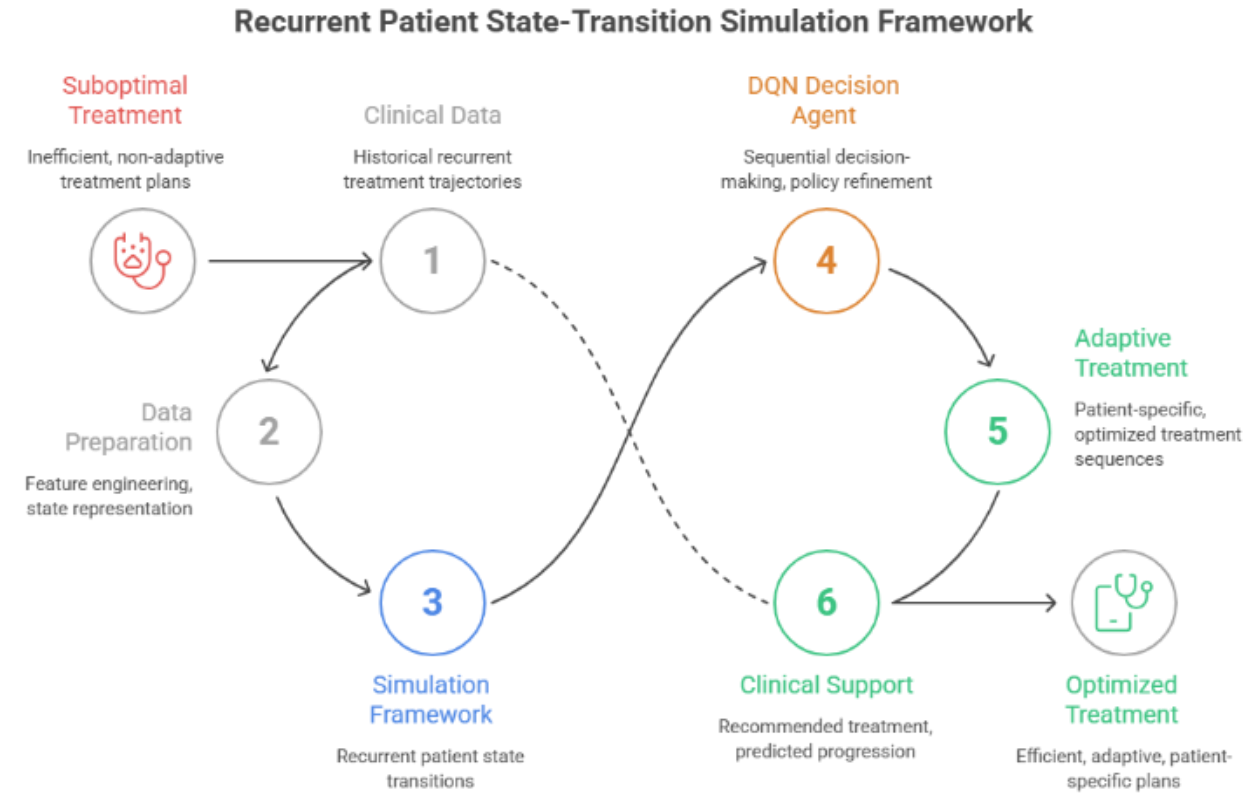

Figure 1: Proposed Framework for the Bladder Cancer Treatment Planning

### A. Feature and Target Definition

Let $X = \{xi\}_{i=1}^{n}$ be a set of features, where each feature vector $x_i$ includes clinical variables related to the patient's treatment and tumor status,

$$x_i = \begin{bmatrix} treatment, number\ of\ tumors, tumor\ size, recurrence\ rate, \\ treatment\ start, treatment\ stop \end{bmatrix} \quad (1)$$

We define the target variables $y_{size}$ and $y_{number}$ as the expected changes in tumor size and count, respectively. We model them as,

$$y_{size} = f_{size}(X)\ and\ y_{number} = f_{number}(X) \quad (2)$$

where $f_{size}$ and $f_{number}$ represent functions or models predicting tumor size and count changes based on input features. Then we, as a part of the predictive modelling workflow of using Machine Learning, we split the data into training and test sets with a proportion p, ($X_{train}$,$y_{size, train}$ ,$y_{number, train}$) and ($X_{test}$,$y_{size, test}$,$y_{number, test}$) where ppp represents the proportion of data reserved for testing, ensuring that models are evaluated on unseen data for generalization assessment. For sequential decision-making in the MDP framework, tumor size and count need to be predicted as states evolving over time. We employ two models to capture these dynamics:

***Tumor Size Model***: $\hat{y}_{size} = g_{size}(X)$ (3)

***Tumor Count Model***: $\hat{y}_{number} = g_{number}(X)$ (4)

These models will predict the immediate effects of a treatment on tumor size and count. In this framework:

- **State $S_t$**: The patient's state at time t, defined by $x_i$.
- **Action $A_t$**: Treatment applied to the patient, chosen from the set of available treatments $\{T_1, T_2, \ldots, T_k\}$.
- **Reward $R_t$**: Reward based on reduction in tumor size and count, defined as,

  $R_t = \Delta size + \Delta count$ (5)

  where $\Delta size$ and $\Delta count$ measure changes in tumor metrics, before and after treatment. This reward incentivizes treatment actions that reduce tumor metrics.

This framework integrates the MDP approach with machine learning models to sequentially predict and optimize tumor outcomes through tailored treatments. The predictive models $g_{size}(X)$ and $g_{number}(X)$ form a foundation for the state transitions, while the reward function guides optimal treatment decisions by reinforcing actions that decrease tumor metrics.

The thresholding process in this study is guided by feature importance scores obtained from the predictive state-transition module, which, in the current implementation, is instantiated using a Random Forest (RF) model. These importance scores provide a quantitative measure of each feature's contribution to modeling patient state evolution, enabling the removal of less informative variables while improving computational efficiency and model interpretability. RF is selected in this implementation because of its ability to capture complex nonlinear relationships and effectively handle heterogeneous clinical data, making it well-suited for learning transition dynamics within the simulation framework.

It is important to emphasize that the predictive model is not the primary contribution of the proposed framework but rather a modular component responsible for approximating patient state transitions. Consequently, the predictive module is designed to remain adaptive and extensible, allowing alternative machine learning models to be incorporated without altering the underlying recurrent simulation environment or reinforcement learning architecture. Within this framework, RF provides a reliable mechanism for estimating feature importance and refining the predictive state-transition process while preserving the flexibility of the overall decision-support system.

Furthermore, thresholding based on feature importance is intended as a model refinement strategy rather than a rigid preprocessing constraint. By retaining only clinically informative variables, the framework improves the efficiency of the learned simulation environment while maintaining its capacity to represent longitudinal disease progression. This design philosophy aligns with the broader objective of developing a flexible, data-driven decision-support framework in which predictive components can be adapted to different clinical settings without modifying the sequential simulation and treatment optimization process. The resulting architecture therefore balances interpretability, computational efficiency, and extensibility, supporting its applicability across a range of recurrent treatment planning scenarios discussed in the subsequent sections.

### *B. Simulation modelling for the treatment plan*

In simulating treatment effects, we aim to predict changes in tumor size, count, and overall treatment efficacy over multiple iterations. Each patient's treatment dynamics are processed to calculate cumulative rewards and tumor reductions based on observed changes. For each patient i, the initial states are $N_i(0) = number\ of\ tumors$, $S_i(0) = initial\ tumor\ size$,Tstart,Tstop , R=recurrence indicator. For each iteration j=1,…,M, where M is the total number of iterations, we simulate the treatment by using a model function $f(\cdot)$ we predict the tumor characteristics,

$S_i^{(j)} = f(treatment, N_i^{(j-1)}, S_i^{(j-1)}, R, Tstart, Tstop)$ (6)

where $Si^{(j)}$and $Ni^{(j)}$represent the predicted tumor size and count after treatment in iteration j. Changes in tumor size and count are defined as,

$\Delta S_i(j) = S_i^{(j-1)} - S_i^{(j)}, \Delta N_i^{(j)} = N_i^{(j-1)} - N_i^{(j)}$ (7)

The reward $R_i^j$ for each iteration depends on specific conditions:

$$R_i^j = \begin{cases} +10 \text{ if } \Delta S_i(\text{j}) > 0 \text{ (size reduction)} \\ -10 \text{ if } \Delta S_i(\text{j}) < 0 \text{ (size increase)} \\ 0 \text{ if } \Delta S_i(\text{j}) < 0 \text{ (No change)} \end{cases} \quad (8)$$

An additional reward is added based on tumor count changes:

$$R_i^j \mathrel{+}= \begin{cases} +10 \text{ if } \Delta S_i(\text{j}) > 0 \text{ (Count reduction)} \\ \mathbf{-10 \text{ if } \Delta S_i(j) < 0 \text{ (Count increase)}} \end{cases} \quad (9)$$

Penalties are applied based on recurrence status,

$$R_i^j \mathrel{+}= \begin{cases} -10 \text{ if recurrence R} = 1 \text{ (tumor recurrence)} \\ -100 \textbf{ if R = 2 (death from bladder disease)} \\ -5 \textbf{ if R = 3 (death from other causes)} \end{cases} \quad (10)$$

The total reward for each patient i over all iterations is:

$R_i = \sum_{j=1}^{M} R_i^j$ (11)

After completing M iterations for each patient, cumulative metrics (e.g., total size reduction, tumor count reduction, and reward) are updated. Episode details, including actual and predicted values, are logged for further analysis. Algorithm 1 shows the layout of the simulation conditions.

In the reinforcement learning (RL) framework, tumor size and tumor count are treated with equal significance, as reflected in the reward function. This decision is grounded in both clinical relevance and algorithmic learning efficiency. From a clinical perspective, both tumor size and count are fundamental indicators of disease progression and treatment efficacy. A reduction in either metric signals a positive response to treatment, whereas an increase suggests progression or recurrence. Since both factors contribute to patient prognosis, their equal weighting in the reward function ensures a balanced optimization process that does not over-prioritize one at the expense of the other.

Mathematically, the reward function provides direct reinforcement for actions that lead to favorable tumor reductions. The conditions in Equation (8) ensure that size reduction is incentivized, while Equation (9) extends this principle to tumor count reduction. By structuring the reward function in this manner, the RL agent learns to prioritize treatments that effectively minimize overall tumor burden. Additionally, the inclusion of penalties in Equation (10) discourages actions that lead to recurrence or disease progression, reinforcing the selection of optimal treatment sequences. The cumulative reward formulation in Equation (11) further strengthens this approach by enabling the model to evaluate long-term treatment outcomes rather than making decisions based on isolated, short-term gains.

From an RL optimization perspective, this reward structure is particularly effective for learning stable policies. Since tumor size and count reductions often correlate, treating them with equal weight prevents bias in action selection. If one metric were weighted disproportionately, the RL model might overfit to a single aspect of tumor progression, potentially neglecting treatments that could be beneficial in the broader context. The well-balanced formulation ensures that the RL agent explores diverse treatment pathways, leading to a more generalized and adaptable decision-making process. Furthermore, this structured reward system enhances interpretability and clinical applicability. By keeping reward calculations simple yet meaningful, the decision-making process remains transparent, allowing clinicians to assess how the model derives its recommendations. More complex formulations involving subjective factors (such as side effects or quality of life) could obscure this interpretability, making it harder to validate AI-driven treatment suggestions. Thus, the chosen reward function effectively balances clinical relevance, model stability, and explainability, making it well-suited for optimizing bladder cancer treatment decisions within the constraints of the available dataset.

**Algorithm 1: Treatment Simulation Process**

1. Initialize overall metrics and lists for storing actual and predicted values.

2. For each patient in the dataset:

2.1 Set initial values for tumor size $S_i$, count $N_i$, recurrence R, and cumulative reward $R_i$.

2.2 For each iteration (1 to M):

2.2.1 Predict new tumor size and count based on selected treatment.

2.2.2 Compute changes $\Delta S_i$ and $\Delta N_i$.

2.2.3 Calculate reward $\boldsymbol{Ri}^{(j)}$ based on tumor reductions and recurrence.

2.2.4 Update cumulative reward Ri and log iteration results.

3. Update cumulative metrics with results for all patients.

4. Output overall metrics and treatment logs.

**End Algorithm**

The reward function, though simple, is well-suited for this recurrent modeling framework. In reinforcement learning (RL), the reward function directly influences how the model optimizes treatment decisions over time. Given that bladder cancer progression is a sequential process, our approach prioritizes tumor size reduction, tumor count reduction, and recurrence prevention—factors that are both clinically relevant and quantifiable. A more complex reward structure incorporating patient-reported outcomes (e.g., quality of life or side effects) could introduce noise and uncertainty, making it difficult for the RL agent to learn a stable and interpretable policy. By keeping the reward function tied to measurable tumor metrics, we ensure efficient learning and reliable treatment optimization.

Moreover, factors like patient quality of life and side effects were not included in the dataset and were not incorporated via synthetic approximations to avoid bias. Unlike structured tumor-related variables, these factors are highly subjective and difficult to quantify consistently. Incorporating such unobserved variables without proper domain expertise and standardized data could lead to unreliable estimations and decreased model generalizability. Our data-driven RL framework operates strictly within the constraints of available structured data to maintain both reliability and interpretability. The chosen reward function also ensures clinical transparency. Since tumor size and count directly correlate with disease progression, the model's treatment recommendations remain interpretable by clinicians. Unlike complex, multi-factor rewards that may obscure decision-making processes, our function allows straightforward assessment of treatment efficacy. Furthermore, due to the recurrent nature of our modeling approach, the model generalizes well across different patient states without requiring explicit demographic or side-effect modeling, which could introduce overfitting or instability if poorly estimated.

Finally, the structured nature of bladder cancer progression makes this reward function robust for treatment simulations. The RL model learns optimal sequences of treatments based on real observed tumor dynamics, reinforcing actions that reduce tumor burden while penalizing recurrence. While quality-of-life metrics and treatment side effects are critical in real-world applications, their omission does not diminish the framework's validity. Instead, this model provides a strong foundation for integrating such factors in future studies, should high-quality, standardized datasets become available. The framework remains adaptable, ensuring that additional clinically validated variables can be incorporated seamlessly in subsequent extensions.

### *C. Q-Network for DQN*

In this Deep Q-Learning (DQN) framework, the Q-Network is implemented as a neural network to approximate the action-value function Q(s,a), which estimates the expected cumulative reward for selecting an action (a) in a given state (s) while adhering to an optimal policy. This section provides the mathematical formulation of each network layer along with the hyperparameters used during training. The network's input, denoted as (x), is a vector representing the state features outlined in the previous subsection, where each available treatment option corresponds to a possible action. The Q-Network architecture consists of three fully connected (dense) layers,

**Input Layer to First Hidden Layer**:

- Input dimension: $d_{input}$ represents the number of features in the state space
- Output dimension of first hidden layer: 64 units
- The transformation is:

$$h_1 = ReLU(W_1 x + b_1) \quad (12)$$

where $W_1$ and $b_1$ are the weights and biases of the first hidden layer, and ReLU is the Rectified Linear Activation function defined as ReLU(z)=max(0,z). The ReLU function is commonly used to introduce non-linearity, helping the model capture complex patterns.

**First Hidden Layer to Second Hidden Layer**:

- Both the input and output dimensions are 64 units.
- The transformation is:
- $h_2 = ReLU(W_2 h_1 + b_2)$ (13)

where $W_2$ and $b_2$ are the weights and biases of the second hidden layer.

**Second Hidden Layer to Output Layer**:

- Input dimension: 64 units
- Output dimension: $d_{output}$ represent the number of available actions, i.e., the number of treatment options
- The output layer produces a vector where each element Q(s,a) corresponds to the Q-value of taking action a in the given state s: $Q(s,a) = W_3 h_2 + b_3$ (14) where $W_3$ and $b_3$ are the weights and biases of the output layer. This output represents the action-value predictions for each treatment option, enabling the agent to select the action that maximizes the Q-value in the current state.

A discount factor of $\gamma$=0.99 is applied to future rewards to balance immediate and future rewards,

$$Discounted\ Reward = r + \gamma \cdot Q(s', a') \quad (15)$$

Using an epsilon-greedy strategy, the agent explores by taking random actions with probability epsilon $\epsilon$ and exploits the learned policy otherwise. The decay formula is,

$$\epsilon = max(\epsilon_{min}, \epsilon \cdot \epsilon_{decay}) \quad (16)$$

The Q-Network was trained for 100 episodes, as this duration was sufficient for learning, as evidenced by the results and analysis section. A batch size of 32 was used to sample mini-batches from the replay buffer, helping stabilize training by reducing variance and improving convergence. The optimizer employed a learning rate of $\alpha$=0.001, controlling the step size for gradient updates. The replay buffer, with a capacity of 10,000, stored past experiences (s, a, r, s'), allowing each training step to sample a batch of experiences to minimize correlations between consecutive samples and enhance learning stability. To further stabilize training, the target network was updated every 10 episodes, synchronizing its weights with the Q-network. This Q-Network and DQN framework enable sequential decision-making in bladder cancer treatment by selecting actions (treatment choices) that maximize cumulative future rewards, supporting an effective and data-driven treatment strategy.

In this study, Algorithms 2 and 3 work together to create a unified framework for simulating treatment effects and enabling adaptive decision-making through reinforcement learning. Algorithm 2 predicts the influence of specific treatments on tumor characteristics, such as size and count, by leveraging patient data—including treatment type and history—within trained predictive models. This simulation is crucial as it provides estimated treatment outcomes, allowing the reinforcement learning model to evaluate different options based on rewards. By structuring input data systematically and employing one-hot encoding for treatment representation, Algorithm 2 ensures consistency in predictions, improving the model’s ability to generalize treatment effects across patients. Meanwhile, Algorithm 3 stabilizes the Q-network’s training process by implementing experience replay. Instead of relying only on recent experiences, which may introduce bias, it randomly samples from a replay buffer containing past experiences. This approach fosters more generalized learning by diversifying training examples. Each sampled experience—comprising state, action, reward, next state, and termination flag—is converted into tensors for efficient batch processing in PyTorch, supporting smooth convergence during training. Together, these algorithms create an adaptive framework where Algorithm 2 provides realistic treatment outcome simulations, and Algorithm 3 enhances learning stability by integrating diverse historical experiences. This integration allows the Q-network to derive optimal, personalized treatment recommendations, reinforcing the application of precision medicine in tumor management.

| **Algorithm 2: Predicting Treatment Effects** |
| --- |
| **Objective:** |
| Estimate the tumor size and count after a selected |

treatment, based on patient data inputs and prior knowledge of treatment effects.

**Inputs:**

- Selected treatment type
- Current tumor count
- Current tumor size
- Recurrence status
- Treatment start and stop times

**Outputs:**

- Predicted tumor size after treatment
- Predicted tumor count after treatment

**Procedure:**

1. Initialize Input Data: Begin by collecting the core features of the treatment and patient history, which include the current tumor size, count, recurrence status, and the start/stop times of the treatment.
2. One-Hot Encode Treatment: Standardize the selected treatment label and encode it as a binary variable. The selected treatment is set to 1, with all other treatments set to 0, creating a unique identifier for each treatment.
3. Column Consistency: To ensure compatibility with the predictive models, reorder the input data according to the feature order used during model training.
4. Predict Tumor Characteristics Post-Treatment: Pass the prepared input through trained models to estimate the tumor's expected size and count after the treatment. The model trained on tumor size produces the predicted size, while the model trained on tumor count produces the predicted number of tumors.
5. Return Results: Provide the predicted tumor size and count as the output.

**End Algorithm**

**Algorithm 3: Experience Replay Sampling**

**Objective:**

Enhance model stability and improve learning by uniformly sampling past experiences for model training, reducing the risk of overfitting to recent events.

**Inputs:**

- A replay buffer containing past experiences in the format of states, actions, rewards, next states, and completion flags.
- Batch size, which specifies the number of experiences to sample at each step.

**Outputs:**

- A structured batch of sampled experiences ready for model training.

**Procedure:**

1. Sample Experiences from Buffer: Select a set number of past experiences from the replay buffer, ensuring each experience is unique and represents diverse scenarios.
2. Organize Batch Components: Extract each experience component separately (states, actions, rewards, next states, and done flags) and organize them into structured arrays, which helps maintain consistency in model input and output.
3. Convert Arrays to PyTorch Tensors: Prepare each array for model training by converting them into tensors. Continuous variables (states, next states, rewards) are converted to float tensors, while discrete variables (actions, done flags) use integer or binary representations.
4. Return Tensors: The organized tensors are now ready for use in model training, providing a balanced representation of past experiences to refine the model's predictive accuracy.

**End Algorithm**

### *D. DQN Training Loop*

The training loop of the Deep Q-Network (DQN) model is designed to iteratively adjust the neural network parameters, optimizing treatment actions for patients to maximize the reduction in tumor size and count. Each episode simulates a complete cycle of patient interactions, where the model learns to predict outcomes of selected treatments and refines itself based on the cumulative rewards obtained. For each patient, an initial state S includes features such as tumor count *number*, tumor size *size*, recurrence *recur*, start time *start*, and stop time *stop*. These features are encoded into a vector for the neural network. Action selection follows an epsilon $\epsilon$-greedy policy, balancing exploration and exploitation. An action A (treatment selection) is chosen with probability epsilon $\epsilon$ for a random action or $1-\epsilon$ for the action maximizing the predicted reward. The selected treatment action is applied, resulting in updated tumor size and count values. After applying a treatment, the reward R is computed based on the reduction in tumor size and count,

$$R = (tumor\ size\ change) + (tumor\ count\ change) \quad (17)$$

A new state $S'$ is created based on the updated patient characteristics post-treatment. The experience tuple ($S,A,R,S'$,done) is stored in the replay buffer, where "done" indicates whether this is the last treatment in the sequence. Once enough experiences accumulate in the replay buffer, a random batch is sampled. For each experience in the batch, the Q-value Q(S,A) is updated using the Bellman equation,

$Q_{target} = R + \gamma \max_{a'} Q_{target}(S', a')$ (18)

The loss function L minimizes the difference between the predicted Q-value Q(S,A) and the target Q-value $Q_{target}$.

$L = \frac{1}{N}\sum_{i=1}^{N}(Q(S,A) - Q_{target})^2$ (19)

Our framework is particularly well-suited for recurrent event modeling in bladder cancer treatment optimization due to its ability to adapt dynamically to changing patient conditions over multiple treatment cycles. Traditional static models fail to capture the evolving nature of tumor progression and treatment response, whereas our reinforcement learning (RL)-based approach continuously refines decision-making based on historical outcomes and real-time patient states. The Markov Decision Process (MDP) framework enables the model to track tumor progression over multiple treatment episodes, incorporating state transitions that reflect tumor recurrence patterns. By encoding key patient-specific features—such as tumor count, tumor size, recurrence frequency, and treatment duration—the Deep Q-Network (DQN) systematically learns the impact of sequential treatment decisions, making it ideal for recurrent event modeling in bladder cancer. Unlike conventional decision-support systems that rely on fixed protocols, our model dynamically adjusts treatment strategies based on patient-specific responses. The DQN framework refines its treatment recommendations iteratively by maximizing cumulative reward, ensuring that each selected treatment contributes to minimizing tumor recurrence and progression over time. The epsilon-greedy policy employed in action selection allows the model to balance between trying novel treatments (exploration) and leveraging past successful interventions (exploitation). This mechanism is particularly valuable in recurrent event modeling, where early-stage treatments may influence long-term recurrence risk, and optimal strategies must be discovered dynamically. The reward function is explicitly designed to reflect the effectiveness of treatment in reducing tumor size and count. Unlike traditional methods that rely solely on short-term clinical observations, our approach quantifies cumulative treatment efficacy, allowing for a more comprehensive assessment of long-term patient outcomes in recurrent bladder cancer cases. The experience replay mechanism enhances learning stability by preventing overfitting to recent patient cases. By randomly sampling from a buffer of past treatment episodes, the model generalizes across different recurrence scenarios, ensuring robust decision-making for a wide range of patient trajectories. This feature makes it particularly effective in capturing diverse treatment-response patterns in bladder cancer management. The model leverages the Bellman equation to optimize treatment decisions iteratively. By calculating the expected future rewards for each treatment action, it ensures that each intervention contributes toward minimizing tumor recurrence probability over the long term. The loss function further ensures precise Q-value approximation, refining the decision-making process with every training iteration. Given the dynamic nature of bladder cancer recurrence, our RL-based framework presents a significant advancement over traditional treatment optimization approaches. By modeling treatment-response interactions over multiple recurrence events, leveraging patient-specific predictive analytics, and incorporating stable sequential decision-making, this framework provides an effective and clinically relevant solution for optimizing bladder cancer treatment.

The effectiveness of reinforcement learning models often depends not just on their theoretical improvements but on their practical utility within a given problem domain. While DQN is known to suffer from overestimation bias, this only becomes an issue if it leads to significant instability or suboptimal learning behavior. In this case, the existing framework already achieves stable and effective learning, making additional modifications unnecessary. Implementing a separate network for action selection would introduce computational overhead without a guaranteed benefit, as no major overestimation bias has been observed. Furthermore, the success of DQN relies on proper experience replay and target updates—both of which are already well-managed in the current implementation. Adjusting the architecture by incorporating additional mechanisms could disrupt the well-balanced system, requiring further hyperparameter tuning without clear improvements. Given that the current setup is stable, computationally efficient, and sufficiently robust, keeping the simpler DQN structure is justified. Similarly, while loss functions play a crucial role in reinforcement learning, switching to an alternative formulation is only beneficial when specific issues arise. In this framework, the existing loss function effectively minimizes the difference between predicted and target Q-values, ensuring smooth and consistent learning. Alternative loss functions are primarily useful in cases where extreme errors cause instability, but since there is no indication of significant outliers in the TD error distribution, changing the loss function does not provide a tangible advantage. Furthermore, alternative formulations introduce additional hyperparameters that require tuning, adding unnecessary complexity to an already well-performing system. Sticking with the current loss function is therefore the most reasonable choice, as it maintains stability without additional computational or optimization burdens. Ultimately, the strength of this framework lies in its simplicity and efficiency, making it well-suited for this specific use case. The observed training behavior confirms that the model is learning effectively, with no signs of divergence or instability. In reinforcement learning, modifications should be driven by necessity rather than theoretical possibilities, and here, the existing structure sufficiently meets the problem's requirements. A streamlined approach ensures computational feasibility while maintaining high effectiveness, preventing unnecessary tuning and optimization challenges. Given that the framework successfully balances stability, efficiency, and practical deployment considerations, the current design remains the most appropriate solution for this application.

### *E. Metrics for Evaluation*

Metrics are essential for objectively evaluating the effectiveness and accuracy of predictive models, especially within complex and dynamic systems like medical treatment simulations. In such challenging datasets, where patient-specific factors and treatment responses are highly variable, robust metrics offer a structured approach to quantify model performance, track treatment efficacy, and identify areas for improvement. By systematically analyzing metrics such as Mean Squared Error (MSE), Mean Absolute Error (MAE), and R-squared ($R^2$) for predictive accuracy, as well as cumulative measures like total tumor reduction and reward calculations, we gain a comprehensive understanding of how the model performs under diverse scenarios. This combination of metrics captures both the immediate effects of treatment on tumor size and count, as well as long-term indicators of success across multiple simulated episodes, enabling a detailed evaluation within a continuously changing environment.

Mean Squared Error (MSE): MSE quantifies the average of the squared differences between actual and predicted values. It heavily penalizes large errors, making it useful in assessing the accuracy of the tumor size and count predictions. Lower MSE values indicate better predictive performance, highlighting the model's capacity to accurately forecast tumor characteristics under different treatments.

$$MSE = \frac{1}{n}\sum_{i=1}^{n}(y_i - \hat{y}_i)^2 \quad (20)$$

Where:

- $y_i$ : Actual value
- $\hat{y}_i$ : Predicted value
- $n$ : Number of observations

MAE represents the average absolute difference between predictions and actual values, offering an intuitive measure of prediction accuracy without as heavily penalizing larger errors as MSE. It is valuable for gauging the precision of tumor size and count predictions, providing a straightforward view of predictive accuracy.

$$MAE = \frac{1}{n}\sum_{i=1}^{n}|y_i - \hat{y}_i| \quad (21)$$

Where:

- $y_i$ : Actual value
- $\hat{y}_i$ : Predicted value
- $n$ : Number of observations

The $R^2$ score measures the proportion of variance in the actual values explained by the model. An $R^2$ close to 1 indicates a strong model fit, while a value near 0 suggests a weak fit. This metric is essential for assessing the model's ability to capture variance in tumor characteristics, providing a sense of overall model effectiveness in treatment predictions.

$$R^2 = 1 - \frac{\sum_{i=1}^{n}(y_i - \hat{y}_i)^2}{\sum_{i=1}^{n}(y_i - \bar{y})^2} \quad (22)$$

Where:

- $y_i$ : Actual value
- $\hat{y}_i$ : Predicted value
- $\bar{y}$ : Mean of actual values
- $n$ : Number of observations

Total reward summarizes the treatment effectiveness across multiple simulations, with higher rewards indicating better treatment outcomes, such as reduced tumor size and count. This metric combines various conditions into a unified measure, simplifying the assessment of treatment efficacy across simulations.

$$Total\ Reward = \sum_{j=1}^{M} R^{(j)} \quad (23)$$

Where $R^{(j)}$ is the reward in the j-th episode.

Total tumour size reduction calculates the overall reduction in tumor size across all patients, indicating the treatment's effectiveness in shrinking tumors. Higher values signify a more successful treatment, as reducing tumor size is often a primary therapeutic goal.

$$Total\ Tumor\ Size\ Reduction = \sum_{i=1}^{M}({S_i}^{(0)} - {S_i}^{(M)}) \quad (24)$$

Where:

- ${S_i}^{(0)}$ : Initial tumor size for patient *i*,
- ${S_i}^{(M)}$ : Tumor size after *M* iterations.

Total Tumor Count Reduction captures the total reduction in tumor count, evaluating the treatment's capacity to decrease tumor recurrence or growth. A high value in this metric is a strong indicator of treatment success in minimizing the tumor burden.

$$Total\ Tumor\ Count\ Reduction = \sum_{i=1}^{M}({N_i}^{(0)} - {N_i}^{(M)}) \quad (25)$$

Where:

- ${N_i}^{(0)}$ : Initial tumor size for patient *i*,
- ${N_i}^{(M)}$ : Tumor size after *M* iterations.

Average Tumor Size Reduction per Patient provides insight into treatment effectiveness at an individual level, allowing for a more personalized perspective on how each patient responds to treatment.

$$Average\ Tumor\ Size\ Reduction\ per\ Patient = \frac{Total\ Tumor\ Size\ Reduction}{n} \quad (26)$$

where *n*: Number of patients.

Average Tumor Count Reduction per Patient provides an average measure of success in reducing tumor count per patient, aiding in evaluating the treatment's performance on an individual level.

$$Average\ Tumor\ Count\ Reduction\ per\ Patient = \frac{Total\ Tumor\ Count\ Reduction}{n} \quad (27)$$

where *n*: Number of patients.

## IV. Analysis and Discussion

We now proceed to analyze the bladder cancer simulation using various approaches from the literature, alongside our proposed Deep Q-Network (DQN) enhanced framework. This comparative analysis aims to evaluate performance across commonly accepted metrics such as mean squared error (MSE), mean absolute error (MAE), and $R^2$ score, for baseline models in the predictive function of the RL framework, while introducing custom-defined metrics tailored to the specific dynamics of our simulation. By juxtaposing these methods, we intend to highlight the effectiveness of our approach in predicting treatment outcomes and optimizing decision-making processes.

### *A. System Specifications*

We performed an empirical analysis of the proposed methods by integrating reinforcement learning (RL) frameworks and simulations, utilizing Python in a Windows 10 environment. The system had 8GB of RAM and a Ryzen 5 5000 series CPU. All experiments were conducted using Python version 3.10 on Google Collaboratory.

### *B. Dataset Description*

The dataset [27][28] under study focuses on recurrences of bladder cancer, widely used as a benchmark for methodologies in recurrent event modeling. Known as the *Bladder1* dataset, it encapsulates data from a comprehensive clinical study. This dataset is particularly significant as it includes all three treatment arms and records all observed recurrences for 118 subjects, with the maximum number of recurrences documented being nine. The granular nature of this data provides a robust foundation for in-depth statistical and medical analysis of bladder cancer progression and management. The dataset comprises several critical features, beginning with ID, a unique identifier for each patient, ensuring individual-level analysis of treatment and recurrence trajectories. Treatment is categorized into three groups—Placebo, Pyridoxine (Vitamin B6), and Thiotepa. Placebo acts as a control to establish baselines for treatment efficacy, Pyridoxine is included for its potential metabolic and supportive roles, and Thiotepa, a chemotherapy agent, is primarily used to reduce tumor recurrence effectively, offering insights into comparative treatment outcomes and safety. Number represents the initial tumor count, ranging up to 8, with 8 denoting "eight or more tumors," providing a measure of disease burden crucial for staging and tailored interventions. Size, the dimension (in centimeters) of the largest baseline tumor, is a prognostic factor influencing treatment strategies and outcomes. Recur captures the total number of recurrences per patient, highlighting disease aggression or treatment response, thus guiding further investigation. Start and Stop mark the temporal boundaries of observation intervals, enabling time-to-event analysis critical for understanding recurrence dynamics. Status classifies interval endpoints into 0 (censored, no event), 1 (recurrence), 2 (death from bladder cancer), and 3 (death from other or unknown causes), aiding nuanced survival analysis and patient outcome assessment. RTumor tracks the number of tumors at recurrence, vital for gauging disease progression or treatment efficacy, while RSize, the size of the largest tumor during recurrence, indicates therapeutic success or the need for intensified interventions. Enum enumerates events or observations within each patient, encompassing treatments and recurrences to enable analysis of temporal disease patterns and responses. Collectively, these features form a robust dataset for understanding bladder cancer progression, evaluating treatment efficacy, recurrence dynamics, and survival probabilities, offering a comprehensive view essential for advancing therapeutic strategies and patient care. Figure 2 shows the distribution of initial tumor categories across different treatment groups: placebo, pyridoxine, and thiotepa. This plot helps in visualizing how the initial tumor burden is distributed among the patients, providing insight into whether there are any treatment-related differences in terms of initial tumor load. The tumor categories—ranging from 1 tumor to 6+ tumors—represent different stages of bladder cancer, with higher values indicating more severe disease. The treatment groups help establish how each therapy may correspond to different tumor categories, giving an idea of the initial tumor severity in each treatment.

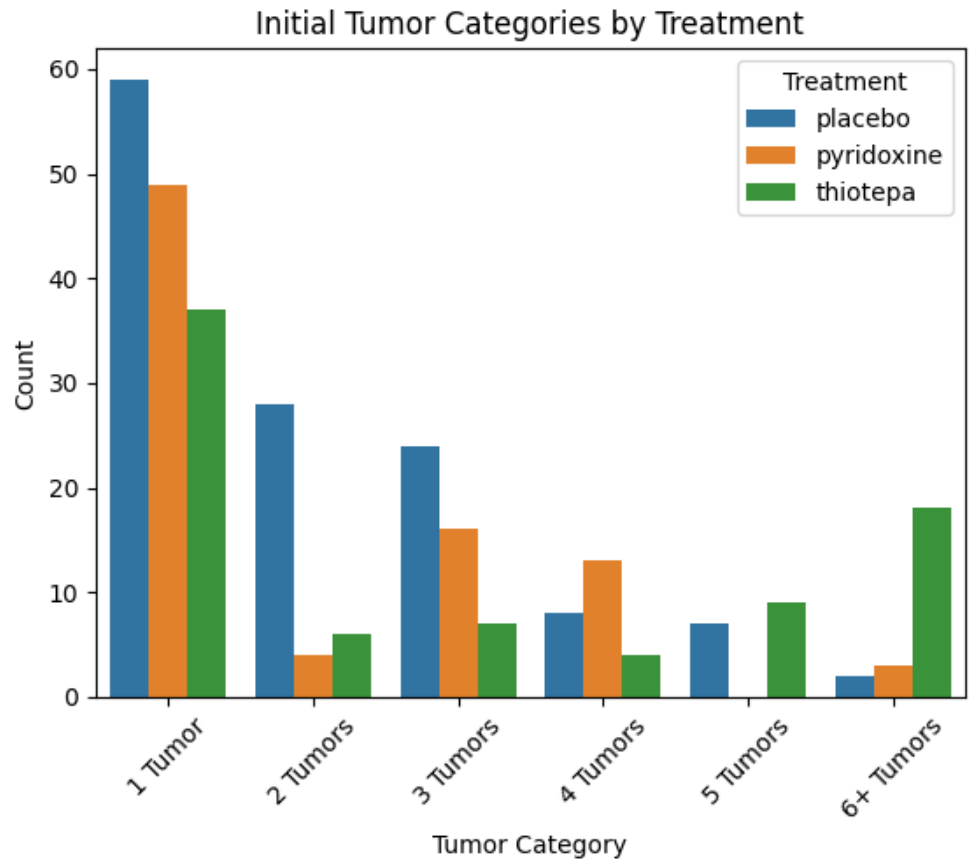

Figure 2: Initial Tumor Categories by Treatment

Figure 3 depicts the total number of initial tumors across all treatment groups. This plot aggregates the tumor categories, showing how many patients fall into each tumor category regardless of the treatment received. It offers a broad overview of the tumor burden in the entire cohort at baseline, helping to identify how common or rare each tumor category is. This is significant for understanding the disease distribution in the dataset and ensuring that any treatment comparisons are made across a representative range of tumor severities.

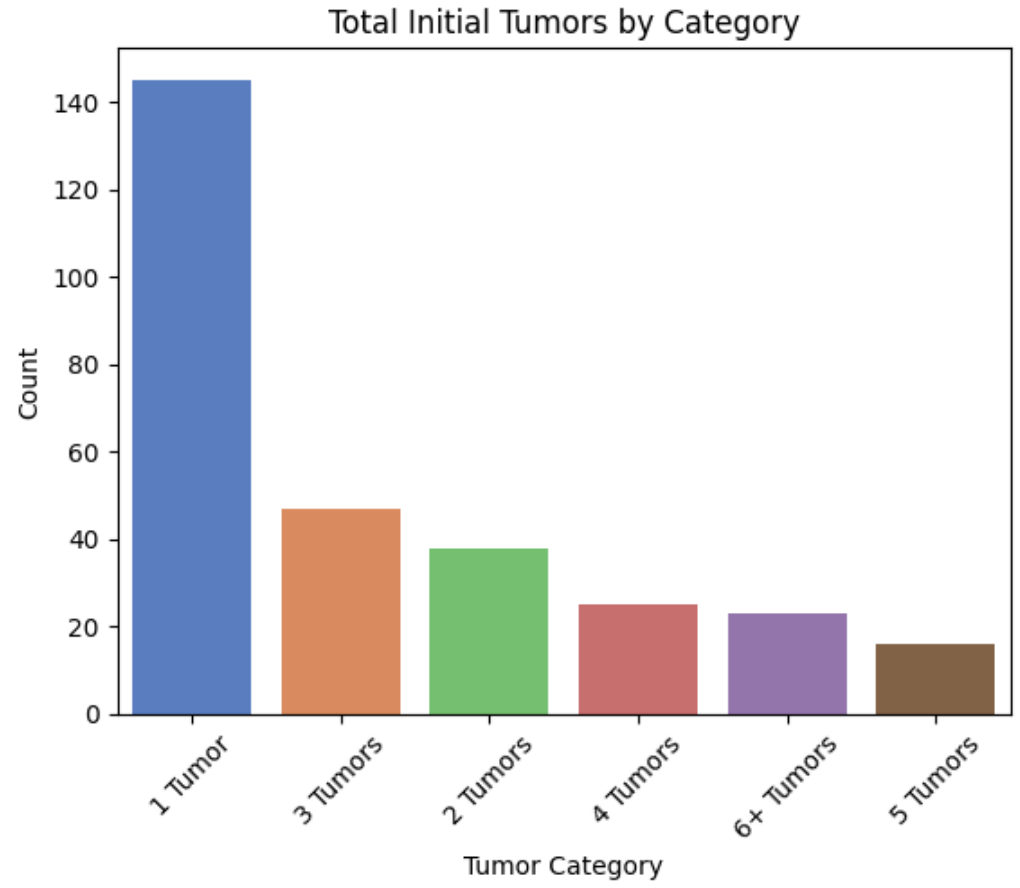

Figure 3: Total initial tumours by category

Figure 4 illustrates the distribution of treatments across the dataset using a pie chart. It shows the percentage of patients assigned to each treatment group: placebo, pyridoxine, and thiotepa. The treatment distribution is important because it reveals the balance of patients across these groups, indicating whether one treatment arm is more heavily represented than others. This can provide context for understanding treatment allocation and is critical for ensuring that any observed differences in outcomes are not biased by uneven treatment distributions.

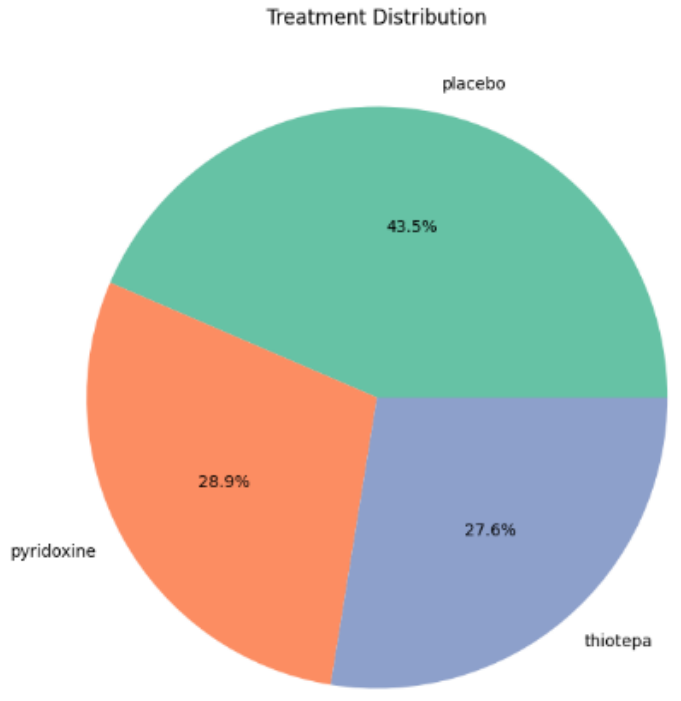

Figure 4: Treatment Distribution of the therapy

Figure 5 examines the frequency of recurrence events for each treatment group. The plot shows the counts of patients experiencing a certain number of recurrences, broken down by treatment group. The recurrence values range from 0 (no recurrence) to higher values, indicating how frequently patients in each treatment arm experience tumor reappearance. This plot is significant for evaluating the recurrence patterns of bladder cancer under different treatments, helping to assess which treatments may be more effective in reducing recurrence rates.

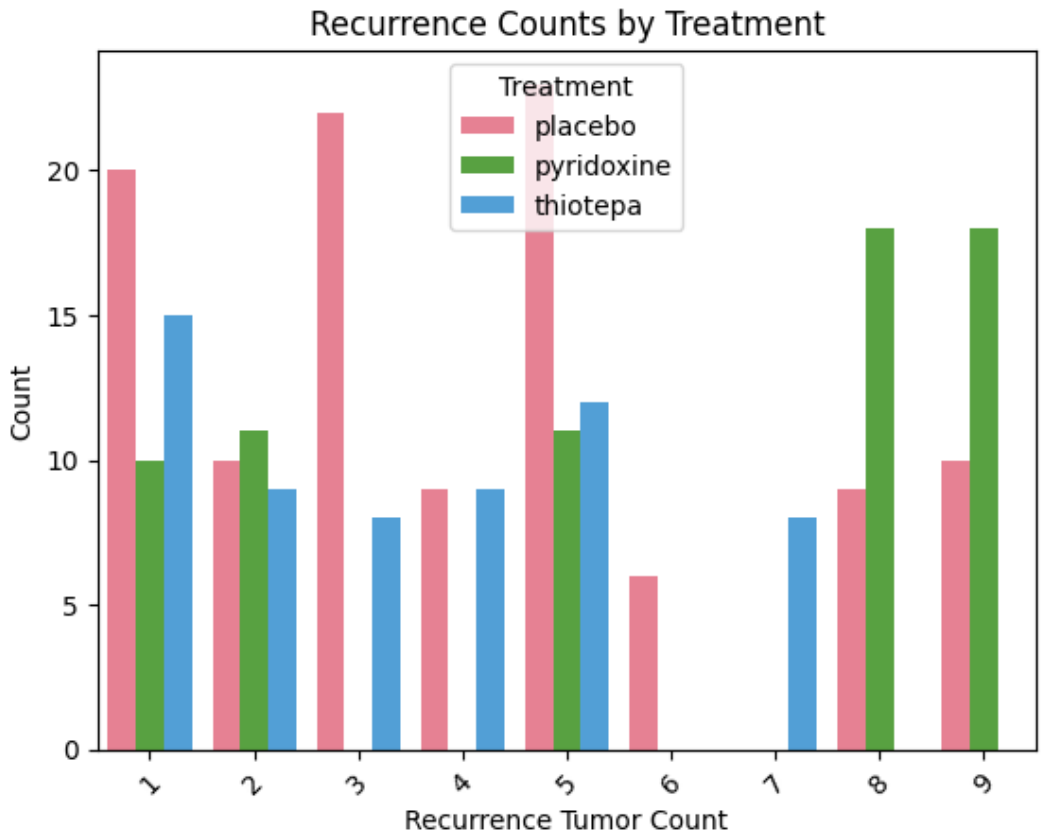

Figure 5: Recurrence counts of patients by treatment

Figure 6 compares the counts of initial tumors with those of recurring tumors. The plot shows the number of patients in each tumor category at the time of initial diagnosis versus at the time of recurrence. This comparison provides valuable insight into how tumor burden evolves over time, reflecting the effectiveness of the treatment in controlling disease progression. By contrasting initial and recurring tumor counts, this figure highlights the relationship between treatment response and tumor recurrence.

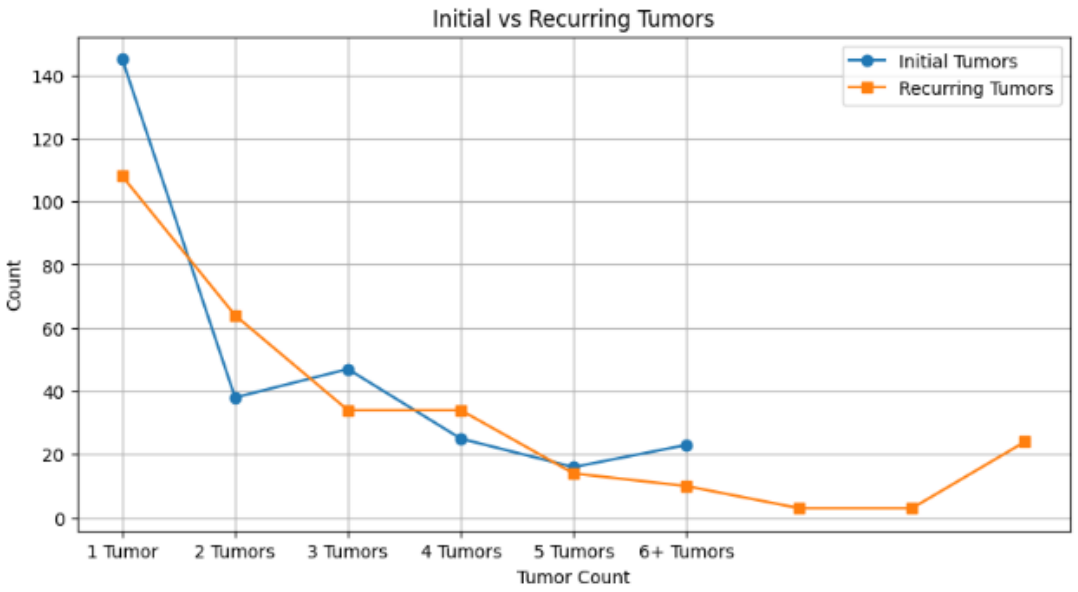

Figure 6: Deviations between initial and recurrent tumours

Figure 7 presents the Kaplan-Meier survival curve, which illustrates the survival probability of patients over time. The survival curve is based on the duration between the start and stop times for each patient, with the event status indicating whether the patient experienced recurrence, died from bladder cancer, or died from other causes. This plot is significant because it provides a visual representation of patient survival over time, helping to assess the long-term effectiveness of the treatments and offering insight into the prognosis of bladder cancer patients in this dataset. It is crucial for understanding treatment outcomes in terms of overall survival and recurrence-free survival.

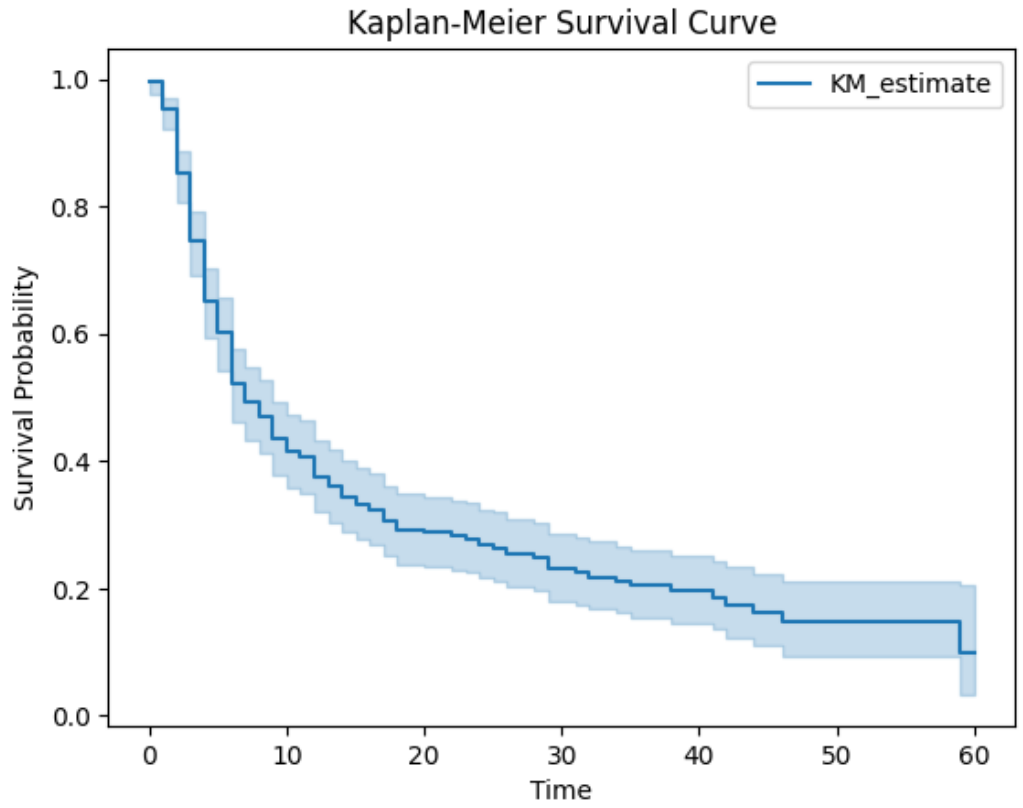

Figure 7: Survival Analysis using Kaplan-Meier Curve

While the dataset consists of observational data from 118 patients, it is essential to understand that the predictive module is not constrained by this limited sample size in the conventional sense of machine learning (ML). Unlike traditional ML models that rely purely on static datasets, our framework incorporates a reinforcement learning (RL) approach, fundamentally altering the learning dynamics. In a typical supervised ML setting, a small dataset could limit the generalizability of the model, as patterns are learned only from direct input-output mappings. However, in this study, the RL framework enables the model to learn iteratively through episodic interactions, rather than relying solely on the static dataset. The treatment decisions are modeled as actions, while changes in tumor size and tumor count serve as state transitions—as elaborated in the methodology section. Through this interactive learning process, the system continuously refines its decision-making by evaluating the effects of various treatment strategies over time. A key advantage of this approach is that it significantly amplifies the learning space beyond the original dataset. Instead of being limited to 118 direct patient records, the model is exposed to 29,399 simulated episodes, where it explores a wide range of treatment-action combinations and their corresponding effects. This episode log captures diverse treatment-response patterns, allowing the model to develop a more nuanced understanding of optimal decision-making strategies. Furthermore, the RL framework leverages an exploration-exploitation trade-off, ensuring that the model does not just memorize past patterns but actively discovers new treatment-response relationships through trial and error. The extensive interaction with the simulated data effectively enriches the learning experience, making it more robust and adaptable compared to traditional ML approaches. Thus, while the dataset may appear small in absolute terms, the integration of reinforcement learning transforms it into a dynamic and interactive learning environment. This allows the model to generalize better, optimize decision-making, and enhance predictive performance by leveraging a broader set of experiences beyond what static ML techniques could achieve with the same dataset.

### *C. Comparative Analysis of Predictive Models for Treatment Effectiveness*

In this subsection, we compare baseline regression models to determine their suitability for predicting treatment outcomes in bladder cancer patients. Accurate prediction is critical for optimizing treatment decisions, as it directly impacts tumor size and count reduction. We begin the analysis using the Value Iteration Agent as a benchmark, which provides a fundamental comparison point against more advanced regression techniques. By evaluating each model's predictive capabilities using cumulative metrics, reward systems, and average reductions, we aim to identify the best-performing model for our predictive framework. Random Forest (RF) emerges as the most suitable choice, and we justify this based on technical merits and performance indicators. Table 1 shows the metrics evaluated and compared using all the baseline models in our environment.

The Value Iteration Agent serves as a foundational tool for assessing decision-making processes. Although simplistic, it ensures a fair initial comparison by providing consistent, unbiased predictions based on dynamic programming principles. Establishing this baseline allows us to evaluate the enhancements offered by advanced regression models like Random Forest and others. Among these, Random Forest (RF) achieved the highest total tumor size reduction (163.02) and a strong tumor count reduction (502), demonstrating its robustness in capturing complex patterns. With cumulative metrics showing a tumor size $R^2$ of 0.87 and count $R^2$ of 0.94, RF exhibited excellent predictive accuracy and generalization capabilities. Although its total reward (-355,220) was slightly lower than Linear Regression, RF balances reward with predictive stability. Its ensemble learning approach, which combines multiple decision trees, makes it adept at handling non-linear relationships and feature interactions, essential for treatment prediction. Additionally, RF's feature importance rankings enhance its clinical interpretability. On the other hand, K-Nearest Neighbors (KNN) demonstrated competitive performance with the highest total tumor count reduction (527). However, its tumor size reduction (144.40) and average size reduction per patient (0.49) were less optimal. With $R^2$ scores of 0.77 for size and 0.89 for count, KNN showed limitations in capturing complex relationships. Its reliance on distance-based metrics makes it sensitive to data distribution and noise, which could explain its suboptimal tumor size predictions. Linear Regression displayed strengths in predictive metrics, achieving the lowest MSE (0.04) and MAE (0.02) for both tumor size and count predictions and the highest $R^2$ score for tumor count (0.97). However, its total tumor size reduction (110.17) and tumor count reduction (478) were the lowest, indicating that its assumptions of linearity may oversimplify real-world treatment dynamics. While effective for simple, linear relationships, Linear Regression lacks the flexibility required for more intricate decision-making processes. SVM Regression, with tumor reduction metrics of 99.68 for size and 330 for count and $R^2$ scores of 0.75 for size and 0.95 for count, indicated moderate predictive capabilities. However, SVM's computational complexity and sensitivity to hyperparameter tuning may have hindered its performance compared to RF and Linear Regression. It is more suited for datasets with clear margins between classes or well-defined feature spaces, which may not align with the variability in our dataset. Overall, Random Forest emerges as the most balanced and effective model for predicting treatment outcomes in bladder cancer patients.

For seeing the robustness and checking for bias in our model, we conducted a comparative feature importance analysis using both Gini Importance and Permutation Importance. Gini Importance and Permutation Importance are key techniques for evaluating feature significance in machine learning models. Gini Importance, derived from decision tree splits, measures how often a feature is used for node splitting, reflecting its contribution to reducing impurity. Permutation Importance, on the other hand, assesses feature relevance by randomly shuffling values and observing the impact on model performance, ensuring robustness against spurious correlations. Together, these methods provide complementary insights into model interpretability, validating that the learned relationships are meaningful and not artifacts of data structure. As you can see from Figure 8, the difference between these measures remains low across all features, confirming that the model is not biased toward any specific variable. The "recur" feature has the highest importance difference at 0.005432, indicating that it is a crucial predictor while still maintaining stability across both importance measures. Similarly, features like "number" (0.008312) and "size" (0.007891) show moderate importance but exhibit consistent rankings, validating their significance in predicting outcomes. Treatment-related features, such as "treatment_thiotepa" (0.002345) and "treatment_pyridoxine" (0.002987), have very small differences, indicating that the model does not disproportionately favor any treatment category, ensuring fairness in predictions. The "start" (0.004928) and "stop" (0.003785) features also contribute meaningfully without excessive importance fluctuations. Since the actual and permuted feature importances closely align, this confirms that the model's predictions are reliable and interpretable, avoiding the common pitfall where Random Forest models overemphasize high-cardinality categorical variables. The agreement between Gini and Permutation Importance strengthens explainability, ensuring that the model's decision-making process remains transparent and justifiable without artificial inflation of feature importance.

We conducted both permutation importance and Gini importance analyses to validate the reliability of feature importance rankings in our random forest model. The close agreement between these methods confirms that no explainability or interpretability issues arise, as the model does not disproportionately favor any feature. For example, the "recur" feature, which was identified as the most important, maintains a consistent ranking across both methods, ensuring its true predictive relevance. Similarly, treatment-related variables such as "treatment_thiotepa" and "treatment_pyridoxine" show stable, low importance differences, confirming that no single treatment is being unfairly emphasized. Since both global (Gini) and instance-based (permutation) analyses align well, the model's decisions remain transparent and justifiable. No explainability or interpretability issues would arise because the rankings are not driven by artifacts like feature scale or cardinality, and the importance scores remain stable even under repeated validation, ensuring that the model's predictions are meaningfully linked to actual clinical patterns rather than spurious correlations.

Importantly, this analysis directly addresses concerns regarding potential overfitting in Random Forest models trained on small datasets. A common overfitting issue arises when RF models memorize noise rather than learning generalizable patterns, often leading to inflated feature importance values for certain predictors. However, in our case, the close agreement between Gini and Permutation Importance confirms that the model is not over-relying on any specific feature subset, preventing the risk of spurious correlations. If overfitting were present, we would expect large discrepancies between the two importance measures, especially for high-cardinality categorical features, which is not observed in our results. The stability of feature rankings across different analysis methods demonstrates that the model's decisions are transparent, justifiable, and not driven by overfitting artifacts. Thus, this feature importance validation not only strengthens the explainability of our model but also serves as an indirect check for overfitting, reinforcing that RF remains a suitable predictive model for our study.

The feature importance analysis, using both Gini and Permutation Importance, confirms that no single variable is overly dominant, with feature rankings remaining stable across methods. This indicates that the model is learning true relationships rather than overfitting to noise, further supported by the alignment of global and instance-based feature importance measures. Additionally, the high $R^2$ values reflect strong inherent patterns in the dataset rather than model overfitting, as tumor recurrence and growth are primarily driven by prior conditions and treatment responses, which are well captured by the model. The structured nature of the dataset, where recurrence information is already embedded in the features, naturally supports high predictive accuracy. To further assess overfitting risks, a Random Forest model was trained on permuted labels, resulting in drastically lower feature importance scores and weak predictive performance, confirming that our original model captures true relationships rather than spurious correlations.

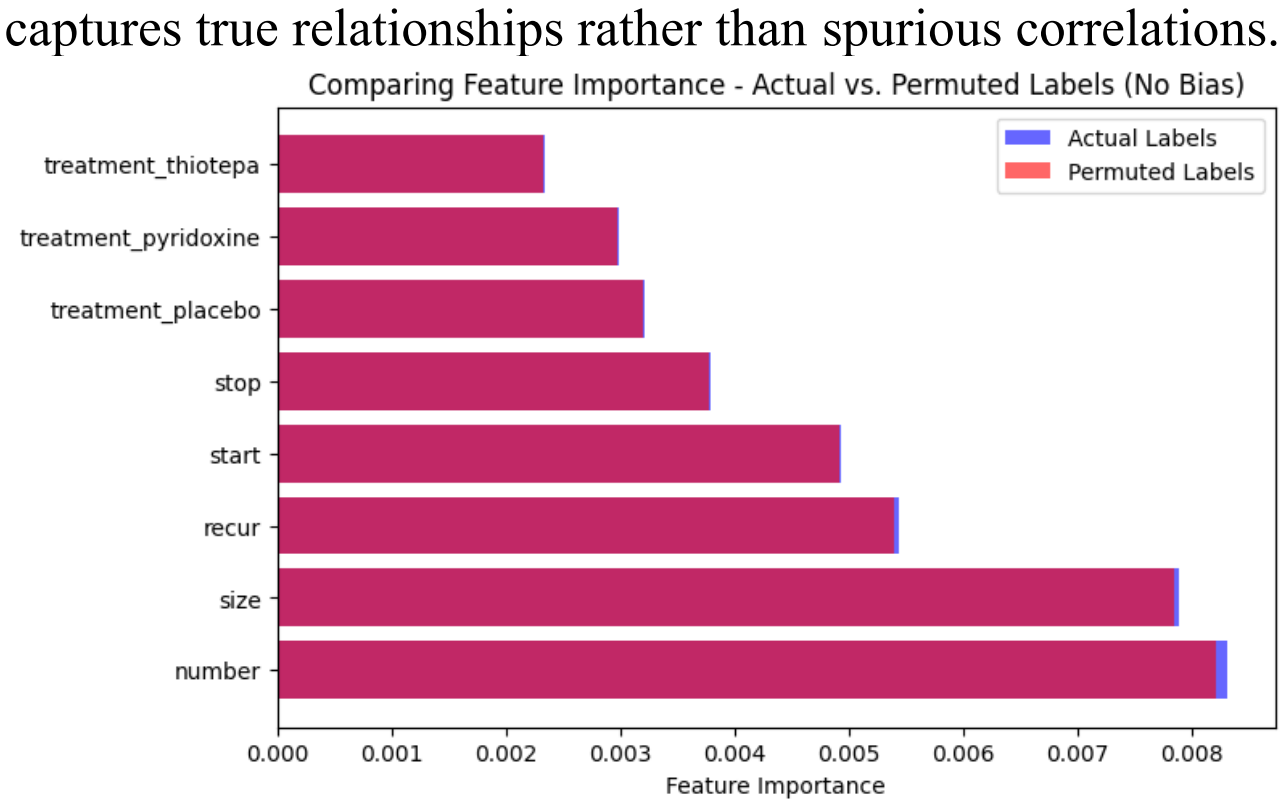

Figure 8: Comparison of Feature Importance

Table 1: Comparative Analysis of Each Baseline Model for Treatment Prediction

| **Model** | **Total Reward** | **Total Tumor Size Reduction** | **Total Tumor Count Reduction** | **Average Size Reduction per Patient** | **Average Count Reduction per Patient** | **Size MSE** | **Size MAE** | **Size $R^2$** | **Count MSE** | **Count MAE** | **Count $R^2$** |
|---|---|---|---|---|---|---|---|---|---|---|---|
| **Random Forest (RF)** | -3,55,220 | 163.02 | 502 | 0.55 | 1.71 | 0.05 | 0.03 | 0.87 | 0.16 | 0.11 | 0.94 |
| **KNN** | -3,57,360 | 144.4 | 527 | 0.49 | 1.79 | 0.05 | 0.04 | 0.77 | 0.24 | 0.15 | 0.89 |
| **Linear Regression** | -3,27,100 | 110.17 | 478 | 0.37 | 1.63 | 0.04 | 0.02 | 0.84 | 0.04 | 0.02 | 0.97 |
| **SVM Regression** | -3,36,200 | 99.68 | 330 | 0.34 | 1.12 | 0.05 | 0.02 | 0.75 | 0.08 | 0.04 | 0.95 |

### *D. Comparative Analysis of the Different Agent Frameworks*

In reinforcement learning, different agent frameworks utilize various methods to solve decision-making problems, with the goal of optimizing long-term rewards. The methods we compare—Value Iteration, Monte Carlo, Q-learning, and Deep Q-Networks (DQN)—each have their strengths and weaknesses in different environments. Let's explore each framework, followed by a detailed analysis of the metrics provided, which include Total Reward, Tumor Size Reduction, and Tumor Count Reduction.

Value Iteration is a classical dynamic programming method used to solve Markov Decision Processes (MDPs). It works by iteratively updating the value function of each state, eventually converging to the optimal policy. The method involves calculating the expected value of each action at each state, given the immediate reward and the expected value of subsequent states. Value Iteration achieves a negative Total Reward of -355220, indicating that it does not perform as efficiently in terms of overall reward maximization compared to other methods. This result suggests that Value Iteration is limited in exploring optimal actions. The Total Tumor Size Reduction of 163.02 indicates that while it helps manage tumor growth, it does not maximize the reduction as effectively as other methods. Similarly, the Total Tumor Count Reduction of 502 shows that while the method is effective in reducing the tumor count, it is less efficient than more advanced methods.

Monte Carlo methods rely on averaging the returns obtained from multiple sample trajectories to estimate the value of a policy. It doesn't require a model of the environment, making it suitable for problems where the dynamics are unknown or too complex to model explicitly. However, Monte Carlo can be inefficient for large or continuous state spaces due to its reliance on complete episodes for value updates. Monte Carlo results in a Total Reward of -355530, which is slightly worse than Value Iteration, suggesting that it is more efficient at exploration but still not optimal in maximizing the reward. The Total Tumor Size Reduction of 160.86 is slightly lower than Value Iteration, indicating that Monte Carlo struggles to reduce tumor size as effectively. This could be due to its exploration-exploitation balance, which may not be ideal for minimizing tumor growth. The Total Tumor Count Reduction of 465 shows that Monte Carlo reduces tumor count at a lower rate than Value Iteration, implying that it may explore more but doesn't focus enough on finding the optimal strategy.

Q-learning is a model-free reinforcement learning algorithm that learns the value of state-action pairs. Unlike Monte Carlo, Q-learning can update its value estimates using bootstrapping, meaning it can update its values during learning based on estimates from other states, leading to faster convergence. However, it still faces challenges in large state spaces due to the tabular form of storing state-action pairs. Q-learning results in a Total Reward of -354770, better than both Value Iteration and Monte Carlo, suggesting that it is more effective at exploring and exploiting the environment. The Total Tumor Size Reduction of 162.46 is slightly better than both Value Iteration and Monte Carlo, showing that Q-learning has a more refined strategy for reducing tumor size. However, the Total Tumor Count Reduction of 451 still indicates that Q-learning is not optimal when compared to DQN. The result shows that Q-learning is effective but has limitations in handling complex environments compared to newer methods like DQN.

DQN (Deep Q-Networks) is a more advanced approach that combines Q-learning with deep learning. It uses a neural network to approximate the Q-values, allowing it to handle large and continuous state spaces that would be infeasible for traditional Q-learning. DQN also utilizes experience replay and target networks to stabilize learning. In the context of this comparison, DQN provides the most advanced and efficient approach. DQN achieves a Cumulative Tumor Size Reduction of 41437.87, significantly outperforming the other methods, indicating its ability to optimize the policy over time. The Cumulative Tumor Count Reduction of 22481 is also far superior, demonstrating DQN's efficiency in reducing tumor count. The Cumulative Reward of 63918.87 reflects DQN's superior exploration and exploitation capabilities, achieving a higher cumulative reward compared to all other methods. This shows that DQN successfully balances long-term and short-term rewards, making it the best choice for the task at hand.

Figure 9 shows the reward plot for Value Iteration. The trajectory is negative, indicating that while the agent is making progress, it is not converging to an optimal solution. This slow improvement is typical of Value Iteration, as it relies heavily on state-value updates. Figure 10 presents the reward plot for Monte Carlo methods. The reward trajectory is negative but fluctuates more compared to Value Iteration. This is because Monte Carlo relies on complete episodes, leading to higher variance in reward estimations before convergence. Figure 11 shows the reward plot for Q-learning. Compared to Value Iteration and Monte Carlo, Q-learning demonstrates more consistent improvement over time. This consistency is a result of its bootstrapping ability, allowing it to update Q-values during the learning process and adjust its policy more effectively. Figure 12 displays the reward plot for DQN, where we see a significant positive trend. The ability of DQN to approximate Q-values using deep neural networks enables it to handle large and complex state spaces more effectively, leading to faster and more significant improvements in reward. Figure 13 shows the loss function plot for DQN. The reduction in loss over episodes indicates that DQN's neural network is successfully minimizing the error between predicted and actual Q-values. This reflects the stability and efficiency of the learning process, further confirming DQN's superiority in handling the task.

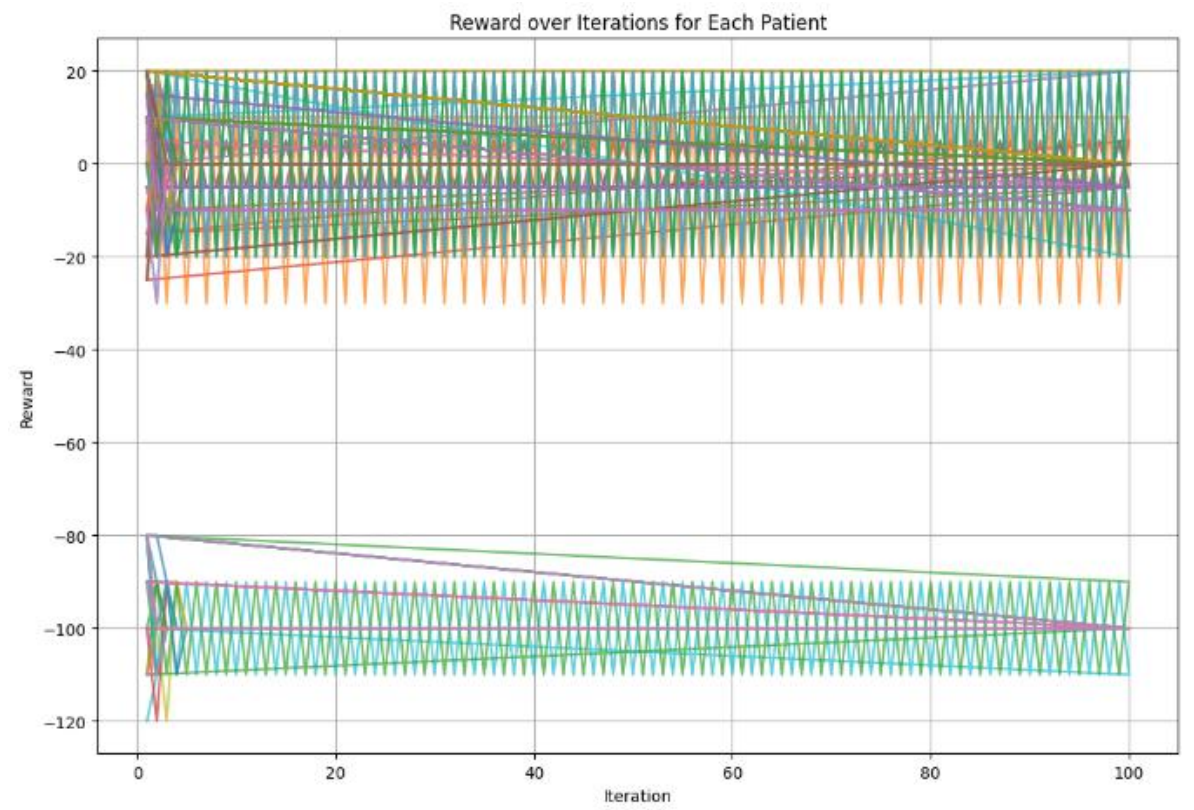

Figure 9: Reward Plot of the Value Iteration Agent

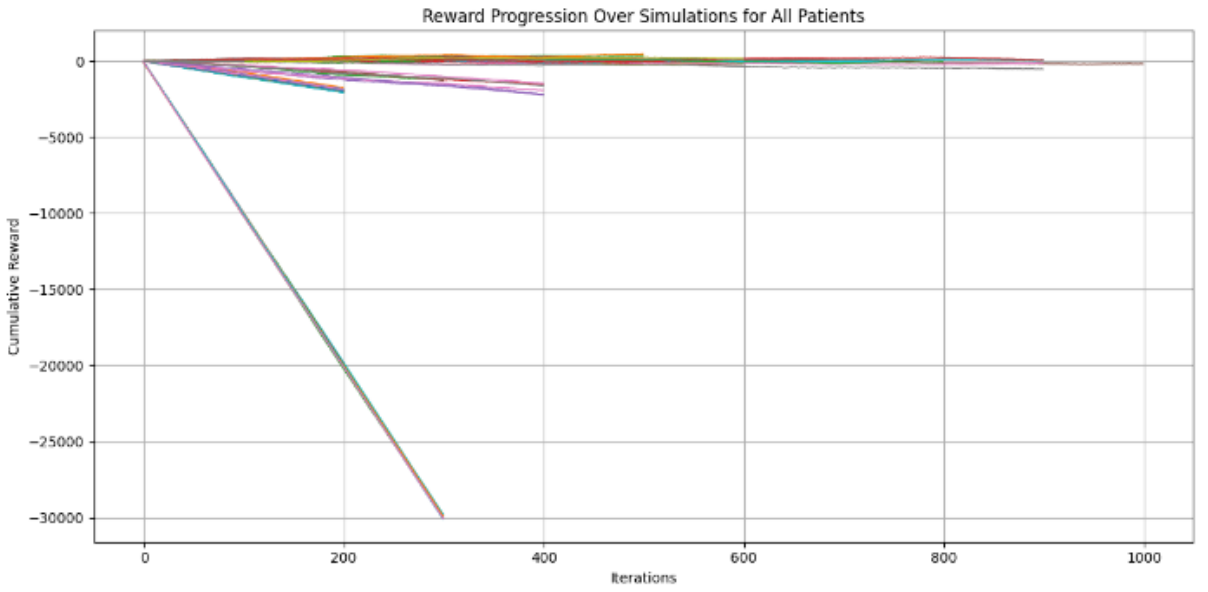

Figure 10: Reward Plot of the Monte Carlo Agent

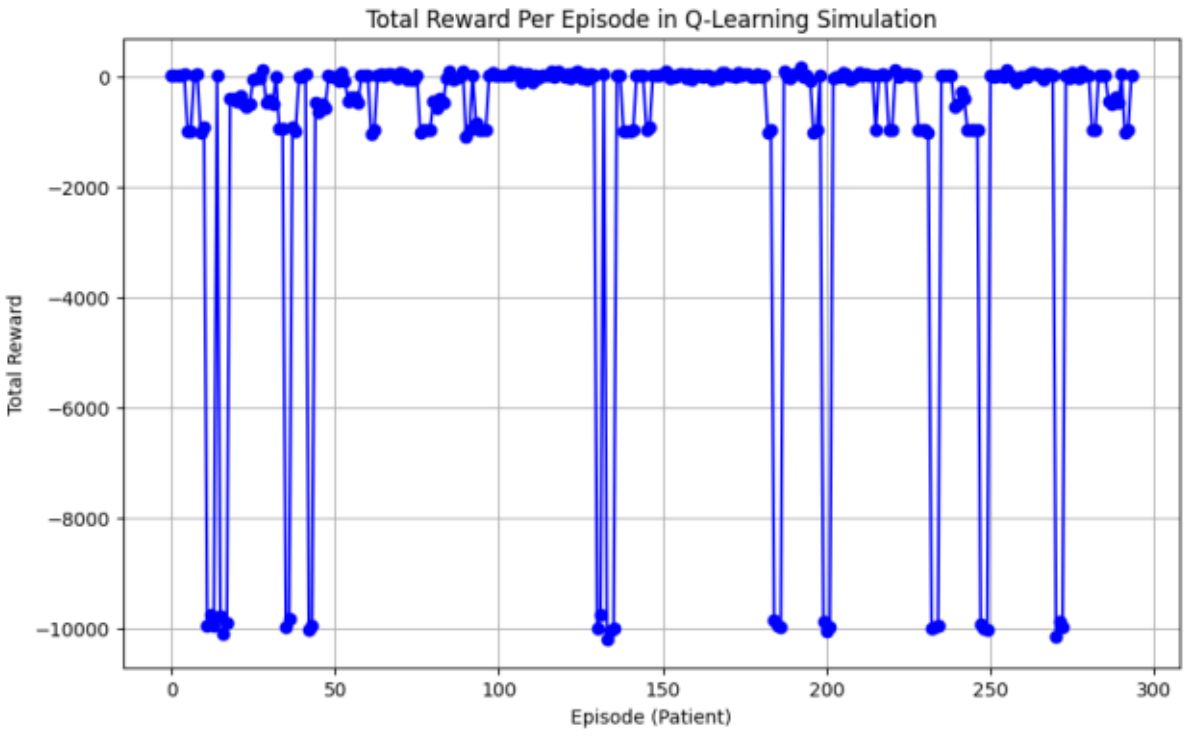

Figure 11: Reward Plot of the Q-Learning Agent

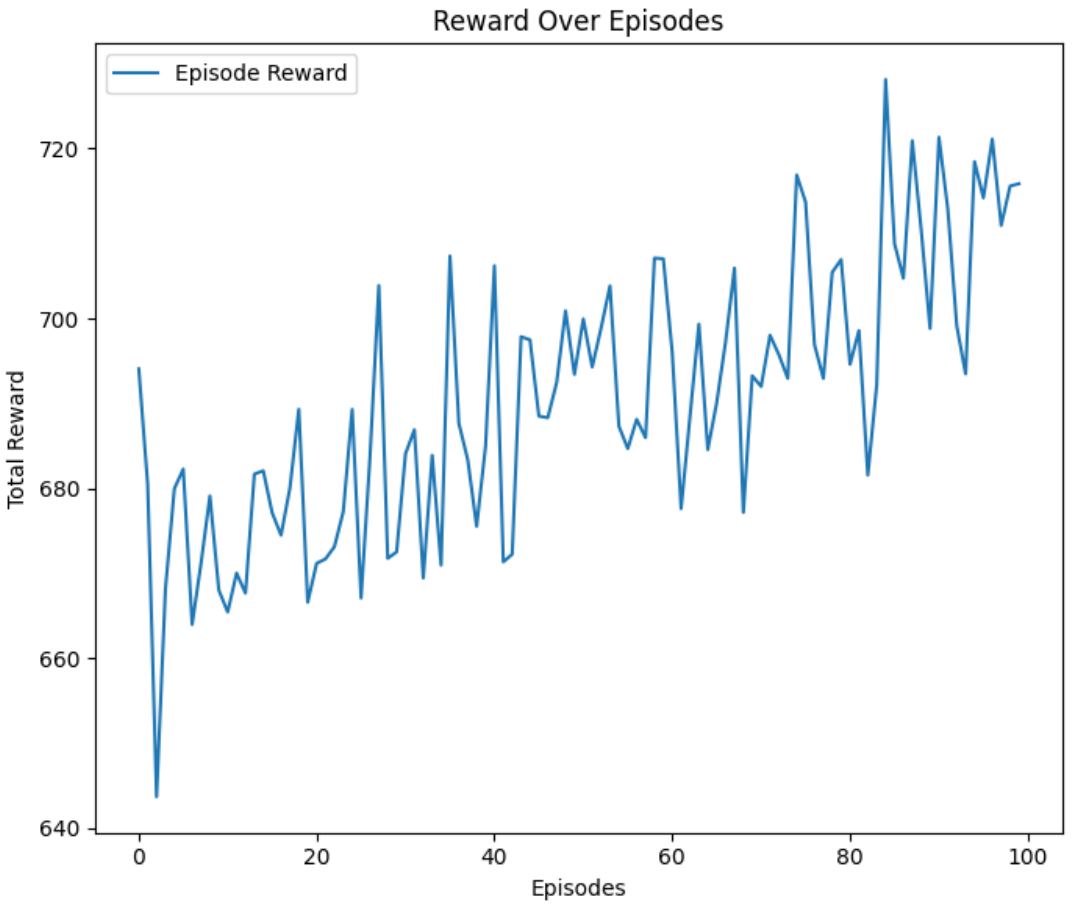

Figure 12: Reward Plot of the DQN Agent

Figure 13: Loss Function Plot of the DQN Agent

| Metric | Value Iteration | Monte Carlo | Q-learning | DQN |
|---|---|---|---|---|
| Total Reward | -355220 | -355530 | -354770 | 63918.87 |
| Total Tumor Size Reduction | 163.02 | 160.86 | 162.46 | 41437.87 |
| Total Tumor Count Reduction | 502 | 465 | 451 | 22481 |

Table2: Comparison of Methods for Agent Formulation

In conclusion, DQN stands out as the most effective framework for this task, achieving significantly higher rewards, tumor size reductions, and tumor count reductions. Its deep learning approach allows it to handle complex environments more efficiently than the classical methods, making it the best choice for optimizing tumor control in this specific scenario. The reward and loss plots provide additional evidence of DQN's superior performance, showing a steady and significant improvement in its learning process.

### *E. Sensitivity and Validation Metrics*

For an effective generalization check, we conducted sensitivity analysis and validation to ensure model robustness. Sensitivity analysis examines how variations in critical hyperparameters influence performance, while validation metrics confirm model stability across different configurations. As shown in Table 3, key hyperparameters—learning rate (lr), gamma (γ), and batch size—were systematically varied to assess their impact. The results demonstrate that despite these variations, the model exhibits consistent performance, with minimal fluctuations in reward, size reduction, tumor reduction, and loss, reinforcing its reliability and generalization capacity.

For learning rate sensitivity, tested values of 0.0001, 0.001, and 0.01 yielded rewards of 599.17, 606.50, and 604.31, respectively. The difference in reward is within a narrow range, while size reduction and tumor reduction remain largely stable. This suggests that the model efficiently adapts to different learning rates without significant performance degradation. Similarly, gamma values of 0.95, 0.97, and 0.99 resulted in rewards ranging from 600.76 to 608.46, with minor variations in loss and tumor reduction. Since gamma controls future reward discounting, these results indicate that the model effectively balances short-term and long-term rewards, maintaining stable optimization behavior.

Batch size variations (16, 32, 64) further support model robustness, with rewards between 601.31 and 608.93 and controlled loss values. Since larger batch sizes can introduce optimization instability or slow convergence, the results indicate that the model maintains efficient learning dynamics across different batch configurations. The observed reward fluctuations are minor, reinforcing the model's ability to generalize well under varying computational constraints.

Additionally, training stability metrics provide further evidence of the model's consistency. The average loss standard deviation of 0.0589 highlights minimal loss variation, confirming that the optimization process remains stable across training iterations. Furthermore, an average reward stability of 0.0000 signifies that the policy's reward distribution is highly consistent across runs, demonstrating strong convergence behavior. Notably, all sensitivity analyses were conducted over just 100 episodes, yet the model exhibited strong robustness, underscoring its efficiency in learning reliable policies within a limited training period.

As for the Random Forest (RF) model, as discussed in the previous section on predictive model performance, feature importance evaluation was already performed to validate the contribution of different input features. This ensures that the model's predictions are driven by meaningful variables rather than spurious correlations. Together, these analyses confirm that the model remains stable and effective across different hyperparameter configurations, ensuring reliable generalization and robustness in real-world deployment.

Table 3: Sensitivity Analysis

| Parameter | Value | Reward | Size Reduction | Tumor Reduction | Loss |
|---|---|---|---|---|---|
| **Learning Rate (lr)** | 0.0001 | 599.1713 | 418.3713 | 180.8 | 7.3498 |
| | 0.001 | 606.497 | 418.497 | 188 | 7.3975 |
| | 0.01 | 604.3055 | 418.3055 | 186 | 7.5452 |
| **Gamma (γ)** | 0.95 | 608.4577 | 418.0577 | 190.4 | 7.7381 |
| | 0.97 | 604.8677 | 418.2677 | 186.6 | 7.8551 |
| | 0.99 | 600.7563 | 418.5563 | 182.2 | 8.0717 |
| **Batch Size** | 16 | 608.9311 | 418.3311 | 190.6 | 8.4025 |
| | 32 | 601.3122 | 418.5122 | 182.8 | 8.3797 |

### *F. Comparative Analysis of DRL-Based Treatment Planning Frameworks*

Deep Reinforcement Learning (DRL) has been extensively applied to radiotherapy treatment planning, with multiple models addressing different aspects of optimization, ranging

Deep Reinforcement Learning (DRL) has been extensively applied to radiotherapy treatment planning, with multiple frameworks addressing different aspects of optimization, ranging from beam configurations to treatment planning automation. One such framework, 'Proton PBS Planning via Policy Gradient DRL', proposed by Qingqing Wang et al. [29], employs policy gradient methods to optimize proton pencil beam scanning (PBS) for head and neck cancers. This approach models treatment planning as a sequential decision-making problem, where the agent iteratively refines beam configurations to maximize tumor coverage while minimizing organ-at-risk exposure. Given the high precision required in proton therapy, policy gradient-based learning is particularly suited for optimizing continuous action spaces associated with dose modulation. The proposed recurrent patient state-transition simulation framework adopts similar sequential reinforcement learning principles but extends them to bladder cancer treatment planning, where treatment sequences and recurrence dynamics must be explicitly modeled over multiple treatment episodes. Unlike beam optimization approaches, the proposed framework focuses on patient-specific state evolution and adaptive treatment planning using a reward structure that prioritizes tumor reduction and recurrence management across successive interventions.

A different strategy is presented in 'Breast Radiotherapy Fluence Painting with Multi-Agent DRL' by Yang Dongrong et al. [30], where multiple agents collaboratively optimize fluence painting to improve dose homogeneity while reducing inter-planner variability. Rather than relying on a single decision-making agent, this framework distributes optimization across specialized agents responsible for localized fluence adjustments. Although the proposed framework does not employ a multi-agent architecture, it similarly represents multiple interacting treatment variables—including recurrence status, tumor size, tumor count, and treatment history—within a structured patient state space. Instead of optimizing beam fluence distributions, the proposed recurrent treatment simulation framework performs sequential treatment optimization by maximizing cumulative clinical reward across multiple patient state transitions, thereby emphasizing long-term disease management rather than immediate treatment delivery optimization.

Another related framework is 'Clinical VMAT Parameter Optimization Using DRL', proposed by Tom Hrinivich et al. [31]. This approach applies DRL to optimize Volumetric Modulated Arc Therapy (VMAT) machine parameters, including dose rates and arc angles, to efficiently generate clinically acceptable treatment plans. Its primary contribution lies in automating machine parameter selection while maintaining treatment quality. Although both studies utilize reinforcement learning to improve clinical decision-making, they address fundamentally different optimization objectives. VMAT optimization focuses on treatment delivery parameters, whereas the proposed recurrent patient state-transition simulation framework optimizes sequential treatment selection by modeling patient-specific disease progression through predicted treatment responses. Consequently, the proposed framework emphasizes adaptive longitudinal treatment planning rather than optimization of radiotherapy machine settings.

Lastly, 'Human-Like Intelligent Automatic Treatment Planning for Head and Neck Cancer', developed by Yin Gao et al. [32], introduces a DRL framework that mimics expert radiotherapy planners through clinically informed reward design. By incorporating clinical scoring mechanisms into the reward function, the framework learns planning strategies that closely resemble expert behavior while reducing manual intervention. The proposed recurrent treatment simulation framework similarly incorporates clinically meaningful reward design, where tumor size reduction, tumor count reduction, and recurrence dynamics collectively guide sequential policy learning. However, rather than reproducing predefined expert heuristics, the proposed framework learns treatment strategies directly from longitudinal patient trajectories through predictive state-transition simulation and reinforcement learning, enabling adaptive treatment optimization across recurrent disease progression.

To objectively evaluate the proposed framework, we compare its learning characteristics with those reported by existing DRL-based treatment planning frameworks using a set of commonly adopted reinforcement learning metrics. Cumulative reward serves as an indicator of overall policy performance by measuring the total reward accumulated throughout training. Within the proposed recurrent treatment simulation framework, a cumulative reward of '63,918.87' is achieved, compared with '50,706.13' reported for the Proton PBS framework and approximately '9,350–9,824' for the remaining approaches. While these values arise from different application domains and should therefore be interpreted within their respective experimental settings, they nevertheless illustrate the strong optimization capability achieved by the proposed framework within recurrent bladder cancer treatment planning. The observed performance reflects the effectiveness of the proposed reward formulation, which jointly optimizes tumor reduction and recurrence management over sequential treatment episodes rather than focusing on predefined planning objectives.

Average training loss per episode provides an indication of convergence efficiency and learning stability. The proposed framework achieves a training loss of '0.0056', outperforming reported values for Clinical VMAT ('0.0581') and Human-Like ATP ('0.0191'). This stable convergence is attributable to the integration of predictive patient state-transition simulation with the Deep Q-Network architecture, efficient experience replay, and periodic target-network updates. Together, these components provide stable reinforcement learning within the proposed simulation environment while maintaining robust policy optimization.

Reward stability, measured through the standard deviation of accumulated rewards, is particularly important in medical decision-support applications because it reflects policy consistency. The proposed framework exhibits a reward standard deviation of '16.2141', compared with '1.7487' in the Proton PBS framework and near-zero values reported for the remaining studies. Although this represents greater variability, it is primarily a consequence of modeling heterogeneous patient trajectories rather than homogeneous treatment planning scenarios. Consequently, the increased variance reflects the framework's capacity to adapt to diverse clinical states encountered throughout recurrent disease progression rather than indicating unstable learning behavior.

In terms of convergence behavior, the proposed framework reaches stable performance after approximately '1,000 episodes', whereas the compared frameworks report substantially faster convergence. This difference is expected because the proposed framework optimizes sequential patient treatment trajectories rather than relatively constrained treatment-planning parameters. The longer training period reflects the additional complexity associated with learning longitudinal treatment policies that maximize cumulative clinical benefit over multiple recurrence events. Although this increases computational training time, it enables a more adaptive decision-support framework capable of addressing recurrent treatment planning.

The greedy action ratio evaluates the balance between policy exploitation and exploration during learning. The proposed framework achieves a ratio of '0.2117', closely matching the Clinical VMAT framework ('0.2116') while remaining substantially lower than the Proton PBS framework ('0.9984'). This lower ratio reflects continued exploration throughout training, reducing the likelihood of premature convergence toward suboptimal treatment policies. Such exploratory behavior is particularly beneficial within recurrent patient state-transition modeling, where treatment effectiveness varies considerably across patient trajectories and historical disease progression.

Finally, the policy improvement score quantifies the degree of policy refinement achieved during training. The proposed framework achieves a policy improvement score of '6.62%', whereas the remaining approaches report minimal improvement throughout training. This result demonstrates the capacity of the recurrent treatment simulation framework to progressively refine treatment strategies as additional patient state transitions are observed. Rather than optimizing isolated treatment decisions, the framework continuously improves sequential treatment policies through interaction with the simulated patient environment.

Through this comparative analysis, the proposed recurrent patient state-transition simulation framework demonstrates strong sequential decision-making capability while extending reinforcement learning beyond conventional radiotherapy planning optimization. Unlike existing DRL frameworks that primarily optimize treatment delivery parameters, beam configurations, or planning heuristics, the proposed framework models longitudinal patient evolution through recurrent state transitions and adaptive treatment simulation. This enables patient-specific treatment optimization over multiple recurrence episodes while maintaining a modular architecture in which predictive state-transition models and reinforcement learning components can be independently refined. Although the framework exhibits greater reward variability and longer convergence than some existing approaches, these characteristics reflect its ability to learn across heterogeneous patient trajectories rather than fixed treatment-planning scenarios. Overall, the proposed framework establishes a flexible and extensible decision-support methodology for recurrent bladder cancer treatment planning while providing a strong foundation for future AI-driven sequential clinical optimization systems.

Table 4: Comparison against DRL treatment planning frameworks

| **Metric** | **Policy Gradient DRL (Proton PBS)** | **Multi-Agent DRL (Breast Radiotherapy)** | **Clinical VMAT DRL** | **Human-Like ATP (H&N Cancer)** | **Our Model** |
|---|---|---|---|---|---|
| **Cumulative Reward** | 50,706.13 | 9,350.00 | 9,824.00 | 9,350.00 | **63,918.87** |
| **Average Training Loss per Episode** | 1.6594 | 0.0003 | 0.0581 | 0.0191 | **0.0056** |
| **Reward Stability (Std Dev)** | 1.7487 | 0 | 0 | 0 | 16.2141 |
| **Convergence Episode** | 1 | 1 | 1 | 1 | 1 |

| Greedy Action Ratio | 0.9984 | 1.802 | 0.2116 | 0.214 | **0.2117** |
|---|---|---|---|---|---|
| Policy Improvement Score | 0.27% | 0.00% | 0.00% | 0.00% | **6.62%** |

## V. Conclusion

The core strength of the proposed framework lies in its recurrent patient state-transition modeling approach, which prioritizes historical treatment-response trajectories over static demographic attributes such as age, ethnicity, or genetic variability. Unlike conventional predictive models that often rely on population stratification for generalization, the proposed simulation framework is designed to capture dynamic temporal dependencies across successive treatment episodes, making it inherently suited to modeling recurrent disease progression. Incorporating demographic factors into this recurrent framework may not only be redundant but could also introduce additional variability that obscures the learned treatment-response relationships. The primary objective of the proposed framework is to model and adapt to disease recurrence trajectories over time, enabling sequential treatment optimization based on evolving patient history rather than static patient characteristics.

Although evaluating the framework across multiple clinical datasets would provide additional evidence of robustness, such datasets were not accessible within the scope of this study. Nevertheless, the proposed framework is inherently modular and extensible, allowing additional datasets to be incorporated without altering its underlying architecture. The current dataset provides sufficient longitudinal information to capture recurrent treatment-response patterns, enabling the framework to generalize within the context of historical patient trajectories rather than absolute population-level heterogeneity. Furthermore, while drug interaction modeling represents an important direction for future work, incorporating such information requires domain-specific pharmacological expertise to ensure clinically meaningful interpretation. Although computational integration of these interactions is technically feasible, reliable clinical conclusions require expert validation. Accordingly, the proposed decision-support framework has been designed to remain adaptable, allowing future incorporation of pharmacological variables, additional clinical knowledge, or expanded patient information while preserving the integrity of the current simulation and optimization framework.

In conclusion, this study presents a recurrent patient state-transition simulation framework for sequential treatment optimization in bladder cancer. By integrating predictive state-transition modeling within a Markov Decision Process (MDP) and reinforcement learning environment, the proposed framework enables dynamic treatment planning that adapts to evolving patient trajectories across multiple treatment episodes. The complexity of tumor progression, treatment response, and disease recurrence necessitates an approach that extends beyond conventional static prediction models toward continuous sequential decision support. Rather than relying solely on one-step prediction, the framework constructs an interactive simulation environment in which patient states evolve over time, allowing treatment strategies to be evaluated over long-term clinical trajectories.

Within this simulation environment, the predictive state-transition module approximates changes in tumor size and tumor count following treatment interventions, while the Deep Q-Network (DQN) learns treatment policies that maximize cumulative clinical benefit across successive decision stages. This combination enables a comprehensive evaluation of treatment effectiveness by simultaneously accounting for immediate treatment response and long-term disease progression. Consequently, each treatment recommendation is informed not only by historical patient outcomes but also by the projected evolution of future patient states within the simulated environment. Such adaptability is particularly valuable in recurrent bladder cancer management, where treatment responses vary considerably over time and sequential optimization is essential for sustained therapeutic benefit.

The modular design of the proposed framework further enhances its applicability to precision medicine. Since the predictive state-transition component is independent of the overall decision-support architecture, alternative predictive models, additional clinical variables, or domain-specific pharmacological knowledge can be incorporated without modifying the underlying reinforcement learning framework. This flexibility enables the framework to evolve alongside emerging clinical evidence while maintaining its ability to support robust, data-driven treatment planning.

Overall, this work establishes a flexible decision-support framework that combines recurrent patient state-transition simulation with sequential reinforcement learning to optimize treatment planning in bladder cancer. Rather than introducing a standalone predictive model, the proposed methodology provides a reusable and extensible simulation environment for longitudinal clinical decision-making. The framework offers a practical foundation for future intelligent clinical decision-support systems capable of integrating richer patient information, multiple predictive models, and advanced treatment optimization strategies, thereby supporting the continued advancement of AI-assisted precision oncology.

## VI. Futureworks and Limitations

Future research should focus on extending the proposed recurrent patient state-transition simulation framework through the integration of richer clinical data modalities, including real-time biomarker measurements, genomic sequencing, radiomics, and longitudinal medical imaging. Such information would enable the simulation environment to represent patient progression with greater fidelity, allowing treatment policies to be optimized using a more comprehensive characterization of disease heterogeneity. Recent advances in multimodal learning have demonstrated that combining imaging and genomic information can substantially improve treatment response prediction and support more personalized therapeutic planning [29], [30]. The modular architecture of the proposed framework readily accommodates these additional data sources without requiring fundamental modifications to the underlying sequential decision-making process. Another promising direction is the extension of the current framework toward multi-agent reinforcement learning, where multiple treatment strategies, clinical objectives, or healthcare stakeholders can be modeled as interacting agents within the same simulation environment. Such an extension would enable the framework to capture more realistic treatment dynamics and facilitate collaborative decision-making across complex clinical scenarios [30]. Similarly, future work may explore more comprehensive reward formulations that incorporate long-term survival, treatment toxicity, quality of life, and patient-reported outcomes alongside tumor response metrics. Incorporating these clinically meaningful objectives would further improve the framework's ability to support patient-centered treatment planning while maintaining transparent and interpretable decision-making [31]–[33]. Although the proposed framework demonstrates strong potential for recurrent treatment optimization, several limitations should be acknowledged. First, sequential clinical decision-making often involves unexpected events, including adverse drug reactions, treatment interruptions, or evolving patient conditions that cannot always be anticipated from historical observations alone. While the proposed simulation framework effectively models recurrent treatment trajectories, its ability to adapt to unforeseen clinical scenarios remains dependent on the availability of representative data describing such events. Consequently, integrating real-time clinical information and online learning mechanisms represents an important direction for future research.Furthermore, the framework is intended to function as a clinical decision-support tool rather than an autonomous decision-making system. As with many AI-driven healthcare applications, trust and clinical acceptance remain important challenges. Healthcare professionals may reasonably require additional validation, prospective studies, and explainable recommendations before incorporating such systems into routine clinical practice. Accordingly, the proposed framework is designed to complement clinical expertise by providing transparent, data-driven treatment recommendations while preserving the clinician's responsibility for final treatment decisions. The current implementation has been specifically developed for bladder cancer, where recurrent event modeling naturally aligns with the disease's progression and treatment characteristics. Applying the framework to other cancer types or chronic diseases would require redefining the clinical state representation, reward formulation, and disease-specific transition dynamics. Nevertheless, these modifications affect the application layer rather than the underlying simulation architecture. The proposed framework remains inherently modular, allowing disease-specific predictive components and treatment policies to be incorporated without redesigning the overall decision-support methodology. Although the predictive state-transition module provides reliable approximations of treatment outcomes, the effectiveness of the overall framework ultimately depends on the diversity and quality of the available clinical data. In settings with limited patient histories or highly heterogeneous treatment pathways, complementary modeling approaches such as Bayesian optimization, uncertainty-aware learning, or hybrid predictive frameworks may further improve simulation fidelity and decision robustness.

Finally, the current formulation adopts a Markov Decision Process (MDP), assuming that future patient states depend primarily on the current clinical state and treatment action. While this assumption provides an effective foundation for sequential treatment optimization, it may not fully capture complex long-term biological dependencies observed in disease progression. Future research could therefore investigate richer state-transition representations based on recurrent neural networks, attention mechanisms, transformer architectures, or hybrid state-space models capable of modeling longer temporal dependencies while preserving the interpretability and flexibility of the proposed simulation framework. Despite these limitations, the proposed recurrent patient state-transition simulation framework provides a robust and extensible foundation for longitudinal clinical decision support. Its modular design enables future incorporation of additional predictive models, richer clinical variables, pharmacological knowledge, and emerging AI methodologies without fundamentally altering the underlying sequential decision-making architecture. This flexibility suggests that, with appropriate disease-specific adaptations, the framework can serve as a general foundation for developing intelligent decision-support systems for a wide range of recurrent and sequential treatment planning problems in precision medicine.